\documentclass[lettersize,journal]{IEEEtran}
\usepackage{amsmath,amsfonts}
\usepackage{algorithmic}
\usepackage{algorithm}
\usepackage{array}
\usepackage{color}
\usepackage[caption=false,font=footnotesize,labelfont=rm,textfont=rm]{subfig}
\usepackage{textcomp}
\usepackage{stfloats}
\usepackage{url}
\usepackage{verbatim}
\usepackage{graphicx}
\usepackage{cite}
\usepackage{booktabs}
\usepackage{multirow}
\usepackage{multicol}
\usepackage{makecell}
\begin{document}

\title{Micro-macro Gaussian Splatting with Enhanced Scalability for Unconstrained Scene Reconstruction}


\author{
  Yihui~Li,
  Chengxin~Lv,
  Hongyu~Yang,
  Di~Huang
}




\maketitle

\begin{abstract}
Reconstructing 3D scenes from unconstrained image collections poses significant challenges due to variations in appearance. In this paper, we propose Scalable Micro-macro Wavelet-based Gaussian Splatting (SMW-GS), a novel method that enhances 3D reconstruction across diverse scales by decomposing scene representations into global, refined, and intrinsic components. SMW-GS incorporates the following innovations: Micro-macro Projection, which enables Gaussian points to sample multi-scale details with improved diversity; and Wavelet-based Sampling, which refines feature representations using frequency-domain information to better capture complex scene appearances.
To achieve scalability, we further propose a large-scale scene promotion strategy, which optimally assigns camera views to scene partitions by maximizing their contributions to Gaussian points, achieving consistent and high-quality reconstructions even in expansive environments.
Extensive experiments demonstrate that SMW-GS significantly outperforms existing methods in both reconstruction quality and scalability, particularly excelling in large-scale urban environments with challenging  illumination variations. Project is available at https://github.com/Kidleyh/SMW-GS.
\end{abstract}

\begin{IEEEkeywords}
Gaussian Splatting, 3D Reconstruction, Unconstrained Scene Reconstruction, Large-scale Scene.
\end{IEEEkeywords}

\section{Introduction}
\IEEEPARstart{3D}{reconstruction} from images is a fundamental and long-standing challenge in computer vision, with widespread applications ranging from immersive virtual reality to 3D content creation. Recent progress in this field has been largely propelled by both implicit representations, such as Neural Radiance Fields (NeRF) \cite{mildenhall2021nerf} and its derivatives \cite{tancik2022block, tpami10716289, tpami10274871}, and explicit representations, exemplified by 3D Gaussian Splatting (3DGS) \cite{kerbl20233d} and related techniques \cite{lu2024scaffold, tpami10965894, tpami10964878, tpami10927644, tpami10870413}. While these approaches have achieved remarkable success in reconstructing static scenes under controlled conditions, which are characterized by stable lighting and consistent camera settings they face notable challenges in real-world scenarios. In unconstrained and dynamic environments, traditional methods often struggle to ensure consistent reconstruction quality, leading to issues such as blurriness, visual artifacts, and significant performance degradation \cite{yang2023cross}.

To address the challenges of dynamic appearance variations in real-world scenes, NeRF-W \cite{martin2021nerf-w} introduced per-image appearance embeddings, which were later refined by methods like \cite{chen2022hallucinated, yang2023cross} to better handle inter-view variations. Despite these advancements, global embeddings often fall short in representing fine-grained details, as they inadequately capture the significant appearance variations influenced by object properties and environmental factors at specific scene locations. Moreover, the high computational demands of implicit representations limit their feasibility for real-time rendering applications.

More recently, Gaussian-based approaches \cite{dahmani2024swag, kulhanek2024wildgaussians} have emphasized modeling the intrinsic features of each Gaussian point, and simultaneously combining the appearance embeddings to predict affine transformations of base colors for dynamic appearance representation through an MLP. While these methods offer improvements, they still encounter limitations akin to NeRF-based approaches, primarily due to the less expressive nature of global embeddings.
The state-of-the-art GS-W method \cite{zhang2024gaussian-w} advances this area by enabling Gaussian points to adaptively sample detailed dynamic appearance information from 2D feature maps, capturing richer details with greater flexibility. However, challenges like blurriness remain, especially upon close inspection of rendered images.

Furthermore, most existing NeRF- and Gaussian-based methods are tailored to single-object scenes, and are not designed to scale to large, complex environments. In large-scale settings, such as urban scenes, significant appearance variations arise across views due to lighting changes, atmospheric conditions, and other environmental factors. These variations frequently introduce inconsistencies in brightness and color, while  supervision for each object in the scene becomes increasingly insufficient. Such complexities highlight the fundamental limitations of current appearance disentanglement techniques, raising critical questions about their ability to generalize to and effectively model the intricate appearance dynamics of large-scale, real-world environments.


In this paper, we introduce Scalable Micro-macro Wavelet-based Gaussian Splatting (SMW-GS), a novel approach that addresses key limitations in existing 3D reconstruction techniques. Our method decomposes Gaussian features into three distinct components: global appearance, refined appearance, and intrinsic features, offering a comprehensive representation of dynamically varying scenes. Global features capture overarching scene attributes, such as color tone and lighting, while refined features model detailed textures and region-specific phenomena like highlights and shadows. Intrinsic features represent consistent characteristics, such as inherent material properties, ensuring robustness across diverse viewpoints and appearance variations.

The core innovation of our work lies in the \textbf{Micro-macro Projection}, which significantly improves refined appearance modeling, which is a critical challenge that existing methods often underperform. Specifically, our approach utilizes adaptive sampling over narrow and broad conical frustums on the 2D feature map, enabling the optimization of 3D Gaussian points to capture both fine-grained textures and broader regional features, such as lighting transitions. 
Inspired by traditional MipMap operations, we introduce a simple yet effective jitter mechanism to the projected position of each Gaussian point at the micro scale, rather than relying on a fixed position as in previous methods. This jitter introduces variability, facilitating the capture of a richer and more diverse set of features.
In addition, we incorporate \textbf{Wavelet-based Sampling}, leveraging frequency domain information to further enhance the accuracy of refined appearance modeling and reconstruction. This multi-scale approach ensures that each Gaussian point effectively captures fine-grained details while preserving feature diversity. To integrate these features cohesively, we design a Hierarchical Residual Fusion Network (HRFN), which seamlessly combines features across different scales, ensuring precise and consistent 3D reconstruction results.


Building on our approach, we next address the challenge of extending Gaussian-level appearance disentanglement to large-scale environments. Conventional large-scene reconstruction pipelines typically adopt a divide-and-conquer strategy to partition scenes into manageable blocks based on visibility or geometric sensitivity, yet they rarely evaluate how effectively those partitions translate into per-Gaussian supervision. To bridge this gap, we introduce our Point-Statistics-Guided (PSG) Camera Partitioning to achieve scene scale promotion. By analyzing the spatial distribution and sampling statistics of each Gaussian point, PSG ensures cameras are assigned where they can maximally inform individual Gaussians, yielding more consistent and robust supervision across the entire scene. Complementing this, our Rotational Block Training scheme alternates between partitioned optimization, stabilizing the appearance disentanglement network under varying supervision distributions. Together, these strategies enable SMW-GS to maintain high geometric fidelity and appearance consistency even when reconstructing expansive, unconstrained urban environments.

Overall, this study makes the following contributions:

\begin{itemize} 
\item Scalable Micro-macro Wavelet-based Gaussian Splatting (SMW-GS): A unified framework for multi-scale 3D reconstruction that decomposes scene representations into global appearance, refined appearance, and intrinsic features to faithfully model dynamic environments. 
\item Micro-macro Projection: A sampling mechanism that combines jittered micro-scale perturbations with adaptive conical frustums, enabling each Gaussian point to capture a diverse range of fine-grained and regional features, significantly improving refined appearance modeling. 
\item Wavelet-based Sampling: A frequency-domain sampling strategy that refines multi-resolution feature representations, enhancing reconstruction fidelity by leveraging high- and low-frequency cues. 
\item Point-Statistics-Guided (PSG) Camera Partitioning: A camera assignment method driven by per-point statistics, which optimally distributes supervision across all Gaussians and ensures balanced training in large-scale scenes. 
\end{itemize}

Extensive experiments on unconstrained image collections, including real-world large-scale datasets and our newly rendered benchmark which is characterized by pronounced appearance variations, demonstrate that SMW-GS consistently outperforms state-of-the-art methods in reconstruction quality.

A preliminary version of this work, MW-GS, was previously published in \cite{li2025micro}. This paper introduces significant improvements over the original version in the following aspects:
(i) Scalability to Large-Scale Scenes: We extend our framework to support arbitrarily large, unconstrained environments by integrating a Point-Statistics-Guided Camera Partitioning strategy that jointly considers block-level sensitivity and per-Gaussian supervision, enabling efficient distributed training and robust appearance disentanglement at scale.
(ii) New Benchmark and Expanded Evaluation: We render a novel large-scale synthetic scene with pronounced appearance variations and, alongside classic in-the-wild and existing large-scale datasets, use them to thoroughly evaluate SMW-GS's generalization and robustness under challenging conditions.
(iii) Fair and Systematic Comparisons: We conduct extensive comparisons against leading in-the-wild reconstruction methods and recent large-scale techniques across multiple datasets, demonstrating clear improvements in reconstruction accuracy, appearance consistency, and overall scalability.

The remainder of this paper is organized as follows. Section \ref{sec:related} reviews related work on in-the-wild and large-scale scene reconstruction. Section \ref{sec:preliminary} provides the necessary background. Section \ref{sec:method} details our proposed method. Section \ref{sec:experiment} presents experimental results and analysis. Finally, Section \ref{sec:conclusion} concludes the paper and outlines future research directions.

\section{Related Work}
\label{sec:related}
\subsection{Scene Representations}
Various 3D representations have been developed to capture the geometric and appearance information of 3D objects or scenes.
Existing traditional methods include meshes \cite{wen2019pixel2mesh, tpami10195242, tpami10226244}, point clouds \cite{qi2017pointnet, qi2017pointnet++, shi2020pv, tpami9547729}, and voxels \cite{schwarz2022voxgraf, wu20153d}. Recently, Neural Radiance Fields (NeRF) \cite{mildenhall2021nerf} have revolutionized the synthesis of novel, photo-realistic views from images. Extensions to NeRF enhance visual quality \cite{barron2022mip, hu2023tri}, rendering speed \cite{chen2023mobilenerf, reiser2023merf}, and convergence \cite{muller2022instant, chen2022tensorf}, though limitations persist in speed and detail.
More recently, 3D Gaussian Splatting (3DGS) \cite{kerbl20233d}, an explicit representation method, offers real-time rendering with high-resolution quality. Recent advances in 3DGS include improvements in efficiency \cite{lee2024compact}, surface reconstruction \cite{huang20242d}, and incorporating semantic attributes for multimodal applications \cite{xie2024physgaussian, shi2024language, qin2024langsplat}. 3DGS has also been extended to various tasks, including autonomous driving \cite{zhou2024drivinggaussian, yan2024street, zhou2024hugs}, 3D generation \cite{chen2024text, chung2023luciddreamer}, and controllable 3D scene editing \cite{chen2024gaussianeditor, wang2024gaussianeditor, zhou2024feature}.

\subsection{Novel View Synthesis from Unconstrained Images}
Traditional novel view synthesis methods assume static geometry, materials, and lighting conditions. However, internet-sourced datasets \cite{snavely2006photo} often contain varying illumination, which challenge these assumptions. NeRF-W \cite{martin2021nerf-w} pioneered addressing these challenges by incorporating learnable appearance embeddings for each image. Later methods like Ha-NeRF \cite{chen2022hallucinated} and CR-NeRF \cite{yang2023cross} further improved appearance modeling using CNN-based encoders.

Despite these advancements, implicit NeRF-based models suffer from slow rendering, leading to the adoption of 3D Gaussian Splatting (3DGS) as a more efficient alternative to NeRF. Approaches such as SWAG \cite{dahmani2024swag} and WildGaussians \cite{kulhanek2024wildgaussians} modulate Gaussian color via MLPs with learnable embeddings. WE-GS \cite{wang2024we} introduces spatial attention for improved CNN-based representations, while GS-W \cite{zhang2024gaussian-w} and Wild-GS \cite{xu2024wild} leverage CNNs to generate feature maps for dynamic appearance modulation. GS-W uses adaptive sampling from projected 2D features, whereas Wild-GS builds triplane embeddings via depth-aware projection.

However, projecting 2D features into 3D often causes sparsity and information loss. In this work, we propose Micro-Macro Wavelet-based Sampling, which enhances sampling diversity and accuracy by incorporating frequency-domain cues, representing the first integration of frequency domain data into 3DGS appearance representation. Our method significantly improves appearance representation and reconstruction quality for unstructured image collections.
\subsection{Large-scale Scene Reconstruction}
Large-scale scene reconstruction has gained increasing attention due to demands for high-fidelity rendering and scalability. Early NeRF-based methods such as Block-NeRF \cite{tancik2022block} and Mega-NeRF \cite{turki2022mega} employ heuristic spatial partitioning. Subsequent approaches like Switch-NeRF \cite{zhenxing2022switch} and Grid-NeRF \cite{xu2023grid} adopt learned or hybrid decompositions.

3D Gaussian Splatting has also scaled to urban scenes via spatial partitioning. VastGaussian \cite{lin2024vastgaussian} and CityGaussian \cite{liu2024citygaussian} adopt a divide-and-conquer approach to reconstruct large-scale scenes. DOGS \cite{chen2024dogs} and Momentum-GS \cite{fan2024momentum} improve training via distributed optimization and self-distillation. However, global appearance embeddings (e.g., in VastGaussian) often struggle to capture fine-grained, per-point appearance changes.
In contrast, our method introduces the first \emph{Gaussian-level} appearance disentanglement within the divide-and-conquer paradigm for large-scale 3DGS. Enabled by Point-Statistics-Guided partitioning and Rotational Block Training, this fine-grained supervision yields superior reconstruction quality and appearance consistency in complex urban environments.


\begin{figure*}[ht]
\centering
\includegraphics[width=0.98\textwidth]{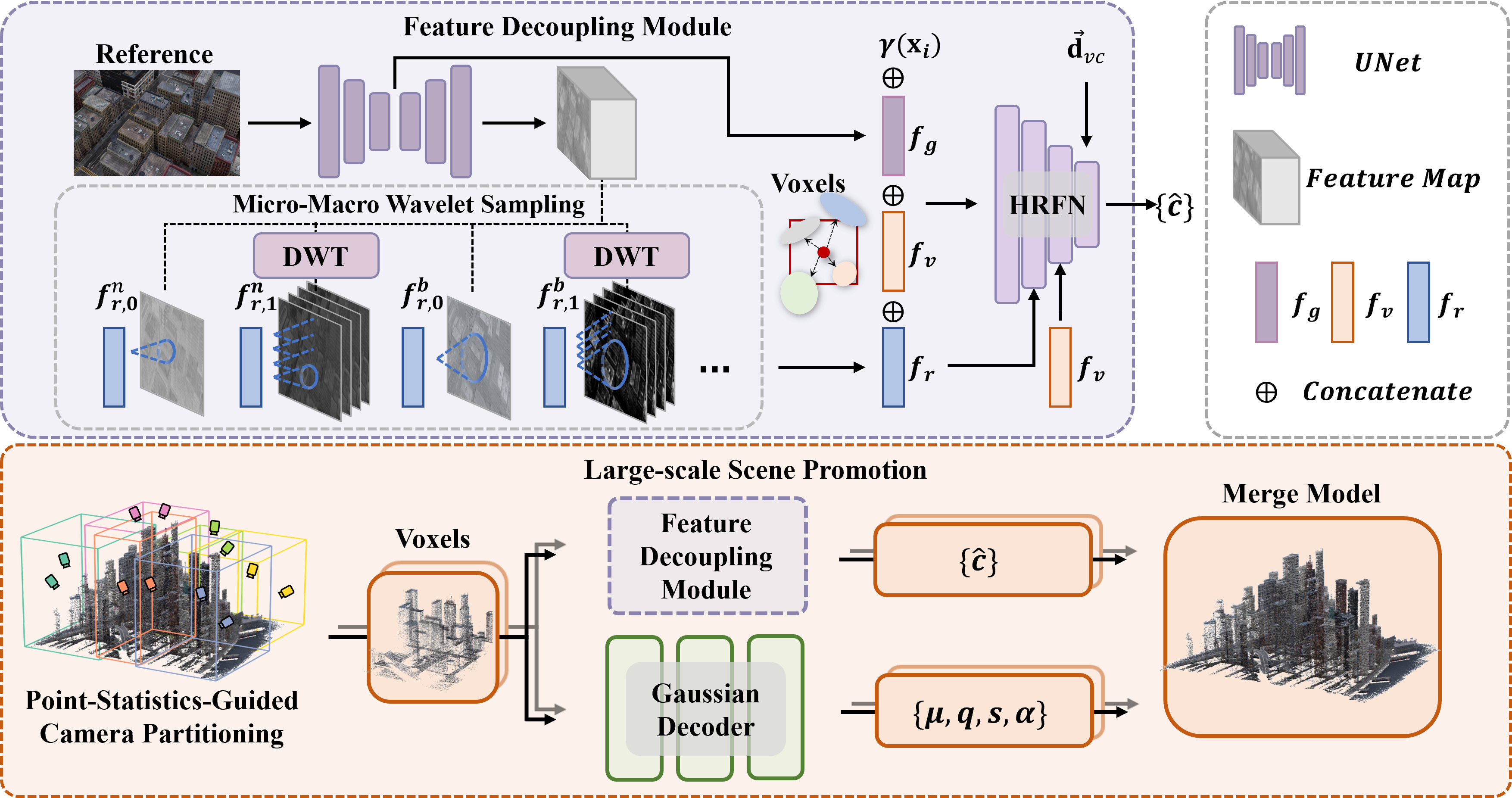}
\caption{Overview of Scalable Micro-Macro Wavelet-based Gaussian Splatting (SMW-GS). Starting from an input image, a CNN backbone extracts global appearance embeddings and multi-scale feature maps. These feature maps undergo a one-level wavelet transform and Micro-Macro sampling, combining jittered micro offsets with broader macro frustums, to capture refined texture details for each 3D Gaussian. The global, refined, and learnable intrinsic embeddings are fused through a Hierarchical Residual Fusion Network (HRFN) to predict per-Gaussian color. For large-scale scenes, Gaussians are organized into overlapping blocks, with camera assignments based on per-point visibility to maximize supervision on the individual Gaussians. An alternating block-wise and full-scene training schedule ensures scalable and consistent reconstruction, supported by a globally unified appearance decoupling module and a shared Gaussian decoder.}
\label{fig:pipeline}
\end{figure*}

\section{Preliminaries}
\label{sec:preliminary}
\subsection{3D Gaussian Splatting}

3D Gaussian Splatting (3DGS) \cite{kerbl20233d} represents scenes using anisotropic 3D Gaussians. Gaussians are projected onto 2D via tile-based rasterization and rendered using fast $\alpha$-blending. Each 3DGS is characterized by a complete 3D covariance matrix $\mathbf{\Sigma}\in \mathbb{R}^{3\times 3}$, which is defined in world space and centered at a point (mean) $\mu\in \mathbb{R}^3$:

\begin{equation}
    G(x) = exp(-\frac{1}{2}(x-\mu)^\top\mathbf{\Sigma}^{-1}(x-\mu)),
\end{equation}
where $x$ is an arbitrary position within the 3D scene. To maintain positive semi-definiteness during optimization, \(\mathbf{\Sigma}\) is decomposed as:
\begin{equation}
    \mathbf{\Sigma} = \mathbf{R}\mathbf{S}\mathbf{S^\top}\mathbf{R^\top}.
\end{equation}

In practice, $\mathbf{R}$ (rotation) is parameterized using a unit quaternion $q$ and $\mathbf{S}$ (scaling) derived from a 3D vector $s$. 
Each Gaussian is further associated with a color $\hat{c}$ and an opacity factor $\alpha$, both modulated by $G(x)$ during blending. For rendering, Gaussians are splatted \cite{zwicker2001ewa} onto the screen, sorted by depth, and employing alpha compositing to compute the final color $\widehat{C}$ for each pixel $\mathbf{p}$:
\begin{equation}
\label{eq:alpha-blending}
    \widehat{C}(\mathbf{p}) = \sum_{i\in N} \hat{c}_i\alpha'_i\prod^{i-1}_{j=1}(1-\alpha'_j),
\end{equation}
where \(\alpha'_i\) is the product of \(\alpha_i\) and the splatted 2D Gaussian contribution.

\subsection{Discrete Wavelet Transform}
Wavelet theory \cite{daubechies1992ten, strang1996wavelets} has long been a foundational tool in image analysis \cite{chen2021local, duan2017sar, li2022wavelet}, offering an effective means to capture both local and global information by describing signals across different frequency bands and resolution levels. 
The 2D Discrete Wavelet Transform (DWT) decomposes an image into four distinct components in the frequency domain using low-pass (\(\mathbf{L}\), emphasizing smooth regions) and high-pass (\(\mathbf{H}\), capturing high-frequency details like textures) filters. Combining these filters yields four unique kernels, namely \(\mathbf{LL}\), \(\mathbf{LH}\), \(\mathbf{HL}\), and \(\mathbf{HH}\), which encode different spatial and frequency information.

Given a feature map \(\mathbf{F} \in \mathbb{R}^{H \times W}\), where \(H\) and \(W\) denote its height and width, respectively, applying a one-level DWT decomposition produces four sub-band features. This process is expressed as:
\begin{equation}
\label{eq:wavelet}
\begin{aligned}
    \mathbf{F}_w^{\mathbf{LL}} &= \mathbf{L}\mathbf{F}\mathbf{L}^\top,
    \mathbf{F}_w^{\mathbf{LH}} &= \mathbf{H}\mathbf{F}\mathbf{L}^\top, \\
    \mathbf{F}_w^{\mathbf{HL}} &= \mathbf{L}\mathbf{F}\mathbf{H}^\top,
    \mathbf{F}_w^{\mathbf{HH}} &= \mathbf{H}\mathbf{F}\mathbf{H}^\top.
\end{aligned}
\end{equation}

For multi-channel feature maps, the wavelet transform is applied independently to each channel. The sub-bands corresponding to the same filter across channels are concatenated, yielding four comprehensive frequency sub-bands that capture diverse spatial and frequency characteristics.


\section{Method}
\label{sec:method}

To address the challenges of reconstructing unconstrained scenes with varying illumination, we propose Scalable Micro-Macro Wavelet-based Gaussian Splatting (SMW-GS), as illustrated in Fig. \ref{fig:pipeline}, a unified framework that enhances 3D Gaussian representations through the following innovations. First, we decompose each Gaussian's appearance into global illumination context, refined multi-scale textures, and intrinsic material embeddings, enabling explicit modeling across different abstraction levels. Global appearance features are extracted from a 2D reference image via a CNN encoder, while intrinsic features are parameterized as learnable embeddings. Our key innovation lies in Micro-Macro Wavelet Sampling mechanism, which enriches refined feature diversity by combining spatial jitter sampling with frequency-domain analysis: at both tight micro offsets and broader macro regions on decoded feature maps, we apply a one-level discrete wavelet transform to capture multi-resolution texture patterns with minimal overhead. A lightweight fusion network seamlessly integrates these signals to predict detail-preserving per-Gaussian color and opacity.  

Crucially, to scale without sacrificing quality, SMW-GS employs a Point Statistics Guided partitioning strategy that dynamically selects camera views for each partition based on per-point visibility statistics. This is paired with a Rotational Block Training scheme that helps maintain uniform optimization of the decoupled module throughout the entire scene, thereby preventing overfitting to local regions. Together, these components guarantee effective supervision for every Gaussian—from isolated objects to expansive urban landscapes, resulting in superior local detail recovery and robust city-scale reconstruction, as corroborated by our extensive experiments.


\subsection{Structured and Explicit Appearance Decoupling}

In unconstrained photo collections, appearance variations stem from factors such as diverse lighting conditions during capture and post-processing operations like gamma correction, exposure adjustment, and tone mapping. Additionally, scene points exhibit directional lighting effects, including highlights and shadows, which dynamically alter their appearance, while intrinsic material properties remain constant.

To systematically model these variations, we explicitly decouple the appearance into three distinct components:
\textbf{Global Appearance Feature} (\(f_g \in \mathbb{R}^{n_g}\)): Encodes overall scene information, capturing coarse-scale lighting and tonal characteristics.
\textbf{Refined Appearance Feature} (\(f_r \in \mathbb{R}^{n_r}\)): Captures detailed, position-specific elements, such as high-frequency textures, local highlights, and shadows.
\textbf{Intrinsic Feature} (\(f_v \in \mathbb{R}^{n_v}\)): Represents the inherent and static properties of scene points.

For a point \(v\) located at \(\mathbf{x}_i\) in 3D space, its appearance is characterized by these three components. The global (\(f_g\)) and refined (\(f_r\)) features are extracted from a reference image, while the intrinsic feature (\(f_v\)) is optimized during training. This structured decoupling balances the global context, local details, and material invariance, providing a comprehensive representation of scene appearance.

To implement this, we adopt a voxel-based organization of Gaussians following the Scaffold-GS framework \cite{lu2024scaffold}. Each anchor point \(v\), located at the center of a voxel, is associated with a scaling factor \(l_v \in \mathbb{R}^3\) and \(k\) learnable offsets \(O_v \in \mathbb{R}^{k \times 3}\), which collectively define the \(k\) Gaussians within the voxel. The global appearance feature (\(f_g\)) is consistently assigned to all anchors within the scene and is derived from a reference image by applying global average pooling to the UNet encoder’s feature map, followed by a trainable MLP (\(MLP^G\)) to produce \(f_g\). This approach ensures consistent modeling of global appearance variations while maintaining flexibility for local and intrinsic attributes.

\begin{figure}[t]    
  \centering        
  \subfloat[GS-W \cite{zhang2024gaussian-w}.]
  {
      \includegraphics[width=0.45\columnwidth]{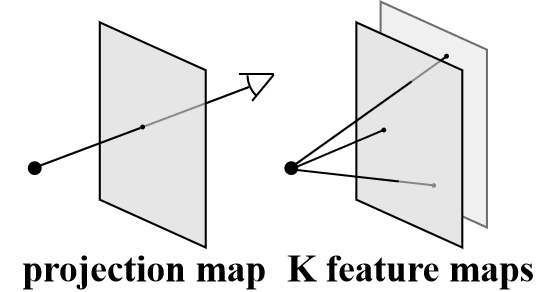}
      \label{fig:gs-w_sample}
  }
  \subfloat[Ours.]
  {
      \includegraphics[width=0.48\columnwidth]{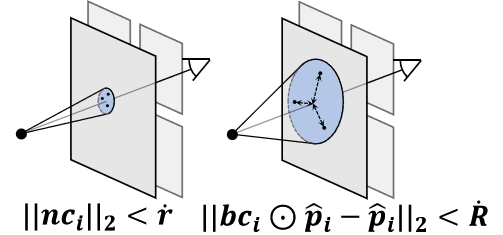}
      \label{fig:our_sample}
  }
  \caption{Sampling comparison between GS-W and our method. The proposed one integrates both narrow and broad conical frustums with wavelet-based sampling, allowing for a more comprehensive capture of features and resulting in enhanced accuracy.}    
  \label{fig:sampling}  
\end{figure}

\subsection{Micro-macro Wavelet-based Sampling}
\label{sec:mws}
To improve the accuracy and richness of scene representation, we propose a novel technique called Micro-Macro Wavelet-based Sampling (MWS). This technique enhances the appearance features of each 3D Gaussian by capturing more detailed and diverse information. It effectively accommodates real-world scene variations by incorporating both fine-grained and broad-scale features. The MWS strategy comprises two main components:

\noindent\textbf{Micro-Macro Projection (MP):}  
Traditional MipMap techniques \cite{williams1983pyramidal} leverage jitter sampling \cite{cook1986stochastic} to introduce random perturbations, enhancing the depiction of texture details. Extending this concept, we propose an adaptive jitter projection method for micro-projections. Instead of directly projecting each 3D point along a ray onto a fixed location on the 2D feature map, our method projects points within a narrow conical frustum. This enables each Gaussian along a ray to capture distinct yet correlated features, reflecting the unique characteristics of each 3D point.

Fig. \ref{fig:gs-w_sample} contrasts our method with GS-W. While GS-W directly projects points onto a \textit{projection feature map}, resulting in identical local appearance features for points along the same ray, it mitigates this limitation by adaptively sampling across multiple feature maps. However, GS-W lacks explicit control over specific local regions, limiting its ability to fully exploit informative features. Our micro-projection method addresses this limitation by employing a narrow conical frustum with a cross-sectional radius parameterized by \( \dot{r} \). To refine the sampling, we introduce \( k_s \) learnable coordinate offsets \(\{nc_i\}_{k_s}\) for each 3D point, enabling adaptive sampling within the frustum. The features obtained from these \( k_s \) samples are averaged to produce the refined feature \( f^n_r \). This design ensures diverse and consistent sampling, capturing rich fine-grained details while preserving texture coherence.

In addition to capturing fine details, MWS also targets broader, long-range characteristics, such as regional  highlights. To achieve this, we employ a broader conical frustum, as shown on the right in Fig. \ref{fig:our_sample}. Guided by the principle that a point’s projection size is inversely proportional to its distance from the camera, we parameterize the projection radius of the broad frustum as \( \dot{R} = \dot{R}_{max} / \| \mathbf{x}_i - \mathbf{x}_c \|_2 \), where \( \mathbf{x}_c \) represents the camera center. We also introduce \( k_s \) learnable scaling factors \(\{bc_i\}_{k_s}\) for each 3D point to enable adaptive sampling within this frustum. The implementation \(bc_i \odot \hat{p}_i\), where \(\hat{p}_i\) denotes the projection center of the frustum, facilitates this process. The features derived from these \( k_s \) samples are averaged to produce the broad appearance feature \( f^b_r \).

By combining the refined \( f^n_r \) and broad \( f^b_r \) features, MWS achieves a balanced representation that captures both intricate details and long-range scene characteristics, significantly enhancing the fidelity and versatility of the scene modeling process.


\noindent\textbf{Wavelet-based Sampling (WS):}  
In unconstrained image collections, significant variation in camera parameters poses challenges in handling large-scale differences across viewpoints with fixed-resolution sampling. To address this, we propose a Wavelet-based Sampling (WS) technique that captures high-frequency and multi-scale information. By leveraging the Discrete Wavelet Transform (DWT), we decompose the feature map \( \mathbf{F}^{MAP} \), generated by a shared UNet with the global feature extractor, into a series of feature maps. The DWT splits \( \mathbf{F}^{MAP} \) into four frequency bands while simultaneously reducing its resolution, effectively preserving spatial information and enabling efficient multi-scale sampling that captures diverse frequency information.

The process begins with dividing the feature map \( \mathbf{F}^{MAP} \) into \(2M+2\) smaller feature maps \(\{\mathbf{F}^1, \cdots, \mathbf{F}^{2M+2}\}\), where each \(\mathbf{F}^i \in \mathbb{R} ^{\frac{n_r}{2M+2} \times H^\mathbf{F} \times W^\mathbf{F}}\). Here, \(M\) is the maximum number of downsampling operations (or the highest DWT level), serving as a critical hyperparameter. The dimensions \( H^\mathbf{F} \) and \( W^\mathbf{F} \) represent the height and width of each smaller feature map, respectively.

During the \(m\)-th downsampling stage, an \(m\)-level DWT is applied to the \((2m+1)\)-th and \((2m+2)\)-th feature maps, producing \( 4^m \) sub-feature maps, as shown in Eq. (\ref{eq:wavelet}). These sub-feature maps are subsequently sampled via bilinear interpolation within narrow and broad conical frustums using the Micro-Macro Projection technique. This yields the feature sets \( \{f^n_{r,m,j}\}_{4^m} \) and \( \{f^b_{r,m,j}\}_{4^m} \) for fine-grained and broad-scale features, respectively.
The refined features \( f^n_{r,m} \) and \( f^b_{r,m} \) for each downsampling level are calculated by applying learnable weight parameters to the sampled features:
\begin{equation}
    f^n_{r,m} = \sum_{j=1}^{4^m} \omega^n_{m,j} \cdot f^n_{r,m,j}, \;
    f^b_{r,m} = \sum_{j=1}^{4^m} \omega^b_{m,j} \cdot f^b_{r,m,j},
\end{equation}
where \( \omega^n_{m,j} \) and \( \omega^b_{m,j} \) denote learnable weights for the \((2m+1)\)-th and \((2m+2)\)-th feature maps, respectively.

Finally, the refined appearance features for each anchor are obtained by concatenating features across all scales:

 \begin{equation}
    f_r = f^n_{r,0} \oplus f^b_{r,0} \oplus\cdots\oplus f^n_{r,M} \oplus f^b_{r,M}.
\end{equation}

By combining Micro-Macro Projection with Wavelet-based Sampling, our method captures multi-scale and high-frequency features, supplementing scene representation with detailed appearance variations and enabling a comprehensive understanding of scene structures across multiple scales.

\subsection{Hierarchical Residual Fusion Network}
\label{sec:hrfn}
To generate the final \( k \) Gaussian colors corresponding to each anchor, it is necessary to effectively combine the global appearance (\( f_g \)), refined appearance (\( f_r \)), intrinsic features (\( f_v \)), and spatial information such as position and view direction. These features exist in different high-dimensional spaces, and simple concatenation is insufficient to achieve effective integration due to the complexity of their interactions. To address this challenge, we propose a Hierarchical Residual Fusion Network (HRFN), which incorporates a hierarchical design with residual connections to enhance the feature fusion process. The HRFN comprises four Multi-Layer Perceptrons (MLPs), denoted as \( \mathcal{M}^H = \{ \mathcal{M}^H_1, \mathcal{M}^H_2, \mathcal{M}^H_3, \mathcal{M}^H_4 \} \).

The inputs to HRFN include the anchor center position \(\mathbf{x}_i\), encoded using a positional encoding function \(\gamma(\cdot)\); the global appearance feature \(f_g\), which encapsulates global information about the scene; the refined appearance feature \(f_r\), which captures multi-scale and high-frequency details; the intrinsic feature \(f_v\), optimized during training to represent specific anchor-level properties; and the direction vector \(\vec{\mathbf{d}}_{ic} = \frac{\mathbf{x}_i - \mathbf{x}_c}{||\mathbf{x}_i - \mathbf{x}_c||_2}\), representing the view direction relative to the anchor. These inputs are processed hierarchically to infer the output colors \(\{\hat{c}_k\}\) for the \( k \) Gaussians. The hierarchical fusion process is formulated as follows:


\begin{equation}
\begin{aligned}
    &Emb = \mathcal{M}^H_1(\gamma(\mathbf{x}_i)\oplus f_v\oplus f_r\oplus f_g) \oplus \omega_r f_r, \\
    &\{\hat{c}_k\} = \mathcal{M}^H_4\left(\mathcal{M}^H_3\left(\mathcal{M}^H_2\left(Emb\right) \oplus \omega_v f_v\right) \oplus \vec{\mathbf{d}}_{ic}\right),
\end{aligned}
\end{equation}
where \(\oplus\) denotes concatenation, and \(\omega_r\) and \(\omega_v\) are learnable adaptive weights that dynamically adjust the contributions of refined and intrinsic features, respectively. 
First, the positional encoding \(\gamma(\mathbf{x}_i)\) is fused with the appearance and intrinsic features through \(\mathcal{M}^H_1\), producing an embedding \(Emb\) that integrates global and local information. A residual term \(\omega_r f_r\) is added to further enhance the representation of refined appearance features. Subsequently, the hierarchical refinement stages (\(\mathcal{M}^H_2, \mathcal{M}^H_3, \mathcal{M}^H_4\)) refine \(Emb\) by progressively integrating the intrinsic feature \(f_v\) and view direction \(\vec{\mathbf{d}}_{ic}\), capturing complex interactions and dependencies among the inputs.

The hierarchical structure of HRFN facilitates a seamless integration of global, refined, and intrinsic features, leveraging their complementarity to capture rich information. Residual connections enhance gradient flow and convergence by preserving original features. This design enables effective modeling of complex feature interactions, leading to accurate Gaussian color prediction.


\subsection{Large-scale Scene Promotion}

Existing in-the-wild methods perform effectively on small, object-centric scenes but struggle in large-scale settings due to their reliance on strong per-Gaussian supervision for Gaussian-level appearance disentanglement. In contrast, large-scale approaches typically adopt block-level supervision, focusing on camera-block relations. This methodological gap leads to misalignment when combining the two, leading subsets of 3D Gaussians receive insufficient guidance. As a result, appearance disentanglement suffers, hindering both Gaussian parameter optimization and U-Net training, and limiting the effectiveness of applying existing divide-and-conquer strategies to large-scale scenes.

\begin{figure}[t]
\centering
\includegraphics[width=0.98\columnwidth]{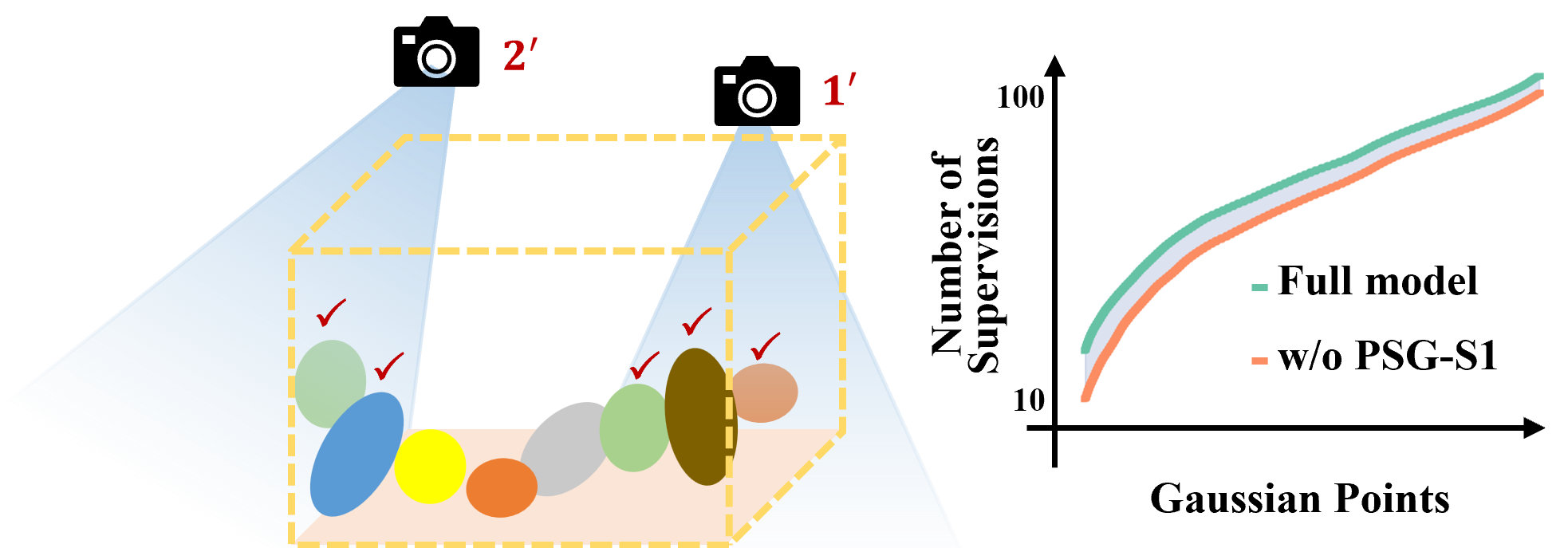}
\caption{Schematic diagram of the Point-Statistics-Guided (PSG) Camera Partitioning in Stage 1. The chart on the right illustrates the variation in the cumulative supervision count for each Gaussian within a block, comparing scenarios with and without the implementation of Stage 1.}
\label{fig:data_partition}
\end{figure}
To overcome these limitations, we focus on enhancing Gaussian-level supervision and adapting our scene representation framework for large-scale environments. We begin by segmenting the scene into blocks using the COLMAP point cloud and then employ a Point-Statistics-Guided camera partitioning strategy. This is further augmented with a block-level camera sensitivity measure to include additional correlated views, ensuring accurate supervision for each Gaussian.
Subsequently, our Rotational Block Training strategy optimizes 3D Gaussians across all blocks for consistent parameter tuning. The appearance disentanglement U-Net leverages the complete image dataset to model appearances with global consistency. By effectively bridging the gap between Gaussian- and block-level supervision, our framework achieves robust, scalable performance, delivering real-time, high-quality rendering in expansive environments.

\noindent\textbf{Initial Division.}
We partition the scene using the COLMAP point cloud by computing the 0.05 and 0.95 quantiles along the ground plane's horizontal axes to remove outliers. The scene is then divided into an $\textbf{M}\times \textbf{N}$ grid of blocks. To ensure overlap, each block's boundary is expanded by 5\% of the corresponding edge length, while outermost blocks are extended infinitely to fully cover all 3D points and camera poses. Cameras and 3D points are initially assigned to blocks based on whether their centers fall strictly within a block's boundaries, serving as the foundation for subsequent partitioning and training.

\noindent\textbf{Point-Statistics-Guided Camera Partitioning.}
To address the insufficient supervision of boundary points in conventional block-based methods, we propose a compound camera partitioning strategy that ensures both minimum supervision guarantees and content-aware camera association. 

Stage 1: Visibility-Aware Camera Allocation (Fig. \ref{fig:data_partition}). We first compute an average visible camera count $\bar{c}$ by projecting each 3D point $p_i$ onto all training views and counting its valid observations. Using a control parameter $\kappa \in (0,1)$, we establish a supervision threshold $\tau = \kappa\bar{c}$. Points with total visible cameras $|\mathcal{V}(p_i)| < \tau$ trigger our compensation mechanism: all cameras observing $p_i$ are directly assigned to its containing block. For remaining cameras, we employ an iterative greedy assignment:
\begin{enumerate}
    \item For each unassigned camera $c_j$, calculate its potential coverage gain $G_j = \sum_{p_k \in \mathcal{B}_m} \mathbb{I}(N_{vis}(p_k) < \tau \wedge c_j \in \mathcal{V}(p_k))$.
    \item Select camera $c_{j^*}$ with maximal $G_j$ and assign to block $\mathcal{B}_m$.
    \item Update $N_{vis}(p_k)$ for all $p_k \in \mathcal{B}_m$ observed by $c_{j^*}$.
    \item Repeat until $\forall p_k \in \mathcal{B}_m, N_{vis}(p_k) \geq \tau$ or no performance gains.
\end{enumerate}
where $\mathcal{B}_m$ denotes the $m$-th block and $\mathcal{V}(p_k)$ the visible camera set of $p_k$.
We present a schematic illustration of Stage 1 in Fig. \ref{fig:data_partition}. The subfigure on the left demonstrates the process of camera selection, while the one on the right depicts the variation in the supervision count for each Gaussian with and without the proposed strategy in Stage 1, highlighting the substantial improvement in supervision effectiveness achieved.

Stage 2: Content-Relevant Camera Augmentation.
Inspired by \cite{liu2024citygaussian, fan2024momentum}, we quantify block-camera relevance through rendering analysis:
\begin{equation}
\Delta \text{SSIM}_j^m = \text{SSIM}(\hat{I}_j, \hat{I}_j^{\setminus m})
\end{equation}
where $\hat{I}_j$ is the full rendering, and $\hat{I}_j^{\setminus m}$ denotes the rendering excluding Gaussian points in block $\mathcal{B}_m$. Cameras with $\Delta \text{SSIM}_j^m > \eta$ (threshold $\eta$) are identified as content-critical and added to $\mathcal{B}_m$'s camera set. This effectively captures cameras whose viewpoints significantly affect $\mathcal{B}_m$'s content.

The combined strategy ensures provable supervision lower bounds through Stage 1's $\tau$-enforced assignment, as well as contextual awareness via Stage 2's rendering-sensitive augmentation, particularly crucial for maintaining consistency in boundary regions where multiple blocks interact.

\noindent\textbf{Rotational Block Training.}
When the number of GPUs matches the total number of blocks (\(\textbf{M} \times \textbf{N}\)), all blocks can be trained in parallel using the full image set. Otherwise, we adopt a \textit{rotational block training} strategy. Blocks are rotated across available GPUs every \(N_{\text{iter}}\), ensuring iterative exposure of the shared U-Net to the entire dataset. This rotational process maintains optimization quality and promotes generalization across all blocks.


\subsection{Training Objective}
Our optimization framework combines photometric supervision with geometric regularization to ensure reconstruction fidelity and physical plausibility. The composite loss function consists of:
\begin{equation}
\mathcal{L}_{photo} = \lambda_{SSIM}\mathcal{L}_{SSIM}(I_r,I_{gt}) + \lambda_1\mathcal{L}_1(I_r, I_{gt})
\end{equation}
for measuring photometric discrepancies between rendered image $I_r$ and ground truth $I_{gt}$. Regularization part contains two components:
$\mathcal{L}_{proj} = \sum\max(\|d_n\|_2-\dot{r},0) + \sum\max(\|d_b\|_2-\dot{R},0)$
constrains projected points within valid frustum regions (Sec. \ref{sec:mws}), where $d_n$ and $d_b$ denote distances to frustum centers in narrow/broad regions, respectively, while $\dot{r}$ and $\dot{R}$ represent predefined distance thresholds for projection constraints; and the volume regularization
$\mathcal{L}_{vol} = \sum_{i} \prod(s_i)$
prevents Gaussian overscaling through scale vector product minimization \cite{lombardi2021mixture,lu2024scaffold}. Here $\prod(\cdot)$ computes the product of Gaussian scale components $s_i$.
The complete objective integrates these terms:

\begin{equation}
\mathcal{L} = \mathcal{L}_{photo} + \lambda_{vol}\mathcal{L}_{vol} + \lambda_{proj}\mathcal{L}_{proj}
\end{equation}

\section{Experiment}
\label{sec:experiment}
To rigorously evaluate the proposed SMW-GS method, we conduct extensive experiments across diverse scenarios. Our evaluation protocol consists of three key components: (1) dataset description, (2) implementation details and evaluation metrics, and (3) comprehensive results analysis with detailed discussions.

\subsection{Datasets and Metrics}
Our experimental evaluation encompasses a diverse collection of datasets, spanning both real-world and synthetic environments. 
For a systematic analysis, we categorize the experiments into three distinct parts: classical unconstrained data reconstruction, large-scale unconstrained scene reconstruction, and synthetic large-scale scene reconstruction under complex appearance conditions. We evaluate all results using PSNR, SSIM \cite{wang2004image}, and LPIPS \cite{zhang2018unreasonable}.

\textit{1) Classical Unconstrained Data Evaluation:} 
We test on three scenes from the Phototourism dataset \cite{snavely2006photo}: \textit{Brandenburg Gate}, \textit{Sacre Coeur}, and \textit{Trevi Fountain}. These datasets contain internet photo collections commonly used for 3D reconstruction. Following prior works \cite{chen2022hallucinated, zhang2024gaussian-w}, all images are downsampled by a factor of 2 for both training and evaluation. Comparisons are conducted against Ha-NeRF \cite{chen2022hallucinated}, CR-NeRF \cite{yang2023cross}, WildGaussians \cite{kulhanek2024wildgaussians}, and GS-W \cite{zhang2024gaussian-w}.

\begin{table*}[ht]
\centering
\caption{Quantitative results on three classical unconstrained datasets. \textbf{Bold} and \underline{underlined} values correspond to the best and the second-best value, respectively. Our method outperforms the previous methods across all datasets on PSNR , SSIM, and LPIPS.}
\label{tab:quantitative-classical}
\begin{tabular}{cccccccccc}
\toprule
\multirow{2}{*}{Method} & \multicolumn{3}{c}{\textit{Brandenburg Gate}} & \multicolumn{3}{c}{\textit{Sacre Coeur}} & \multicolumn{3}{c}{\textit{Trevi Fountain}} \\ 
\cmidrule(lr){2-4} \cmidrule(lr){5-7} \cmidrule(lr){8-10} 
& PSNR $\uparrow$ & SSIM $\uparrow$ & LPIPS $\downarrow$ & PSNR $\uparrow$ & SSIM $\uparrow$ & LPIPS $\downarrow$ & PSNR $\uparrow$ & SSIM $\uparrow$ & LPIPS $\downarrow$ \\
\midrule
Ha-NeRF & 24.04 & 0.887 & 0.139 & 20.02 & 0.801 & 0.171 & 20.18 & 0.690 & 0.223 \\
CR-NeRF & 26.53 & 0.900 &  0.106 & 22.07 & 0.823 & 0.152  & 21.48 & 0.712 &  0.207 \\
WildGaussians &  27.86 &  0.928 & 0.121 &  23.13 & 0.860 & 0.159 & \underline{23.68} &  0.772 & 0.214 \\
GS-W & \underline{27.96} & \underline{0.931} & \underline{0.086} & \underline{23.24} & \underline{0.863} & \underline{0.130} &  22.91 & \underline{0.801} & \underline{0.156} \\
\midrule
Ours &  \textbf{29.37} &  \textbf{0.942} &  \textbf{0.052} &  \textbf{24.64} &  \textbf{0.897} &  \textbf{0.073} &  \textbf{24.07} &  \textbf{0.821} &  \textbf{0.120} \\
\bottomrule
\end{tabular}
\end{table*}



\textit{2) Real Large-Scale Unconstrained Data Evaluation:}
We evaluate the method on four large-scale scenes: \textit{Rubble} and \textit{Building} from the Mill-19 dataset \cite{turki2022mega}, as well as \textit{Sci-Art} and \textit{Residence} from the UrbanScene3D dataset \cite{lin2022capturing}. Consistent with prior works \cite{liu2024citygaussian, fan2024momentum}, all images are downsampled by a factor of 4 during both training and evaluation. The scenes respectively comprise 1,657, 1,920, 2,998, and 2,561 training images, and 21, 20, 21, and 21 test images. We compare the proposed method against state-of-the-art approaches across two categories: \textit{in-the-wild} methods, including GS-W \cite{zhang2024gaussian-w} and WildGaussians \cite{kulhanek2024wildgaussians}, and \textit{large-scale} methods, including VastGaussian \cite{lin2024vastgaussian}, CityGaussian \cite{liu2024citygaussian}, and Momentum-GS \cite{fan2024momentum}.


\begin{figure}[t]
\centering
\includegraphics[width=0.98\columnwidth]{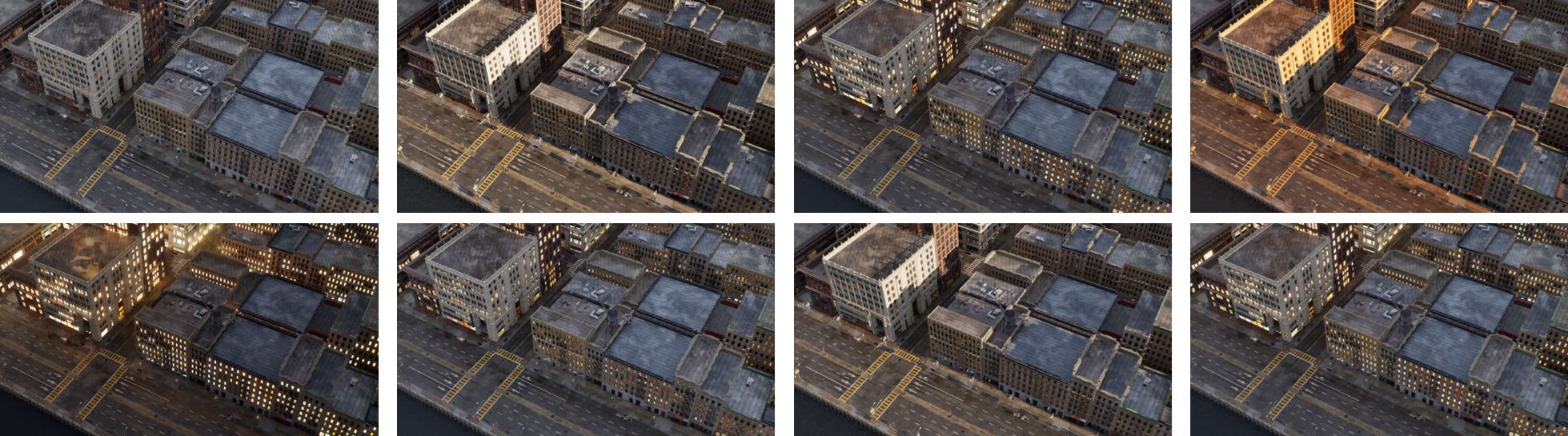}
\caption{An example from our newly rendered MatrixCity dataset showcasing eight distinct appearance conditions.}
\label{fig:matrixcity-8app}
\end{figure}

\textit{3) Synthetic Large-scale Data Evaluation:}
While the \textit{Real Large-Scale Unconstrained Data} provides valuable insights, their appearance variations are limited compared to \textit{Classical Unconstrained Data}. To more thoroughly assess our method's effectiveness in reconstructing large-scale scenes under severe appearance variations as well as fully consistent conditions, we use the synthetic Aerial Data of MatrixCity \cite{li2023matrixcity}, built on Unreal Engine 5. We extend MatrixCity by rendering images under seven additional appearance conditions using Unreal Engine 5 (Fig. \ref{fig:matrixcity-8app}), resulting in eight diverse visual domains. The training set is sampled across all conditions, while the test set follows the original benchmark.

We evaluate performance on both original blocks  (\textit{Block\_A} and \textit{Block\_E}) and newly generated blocks (\textit{Block\_A$\ast$} and \textit{Block\_E$\ast$}), containing 1,063 and 837 training images, and 163 and 124 test images, respectively. All images are used at full resolution ($1,920 \times 1,080$) without downsampling. Metrics and baselines align with those in the \textit{Real Large-Scale Unconstrained Data Evaluation}, ensuring fair and consistent comparison across varying appearance complexities.

\begin{figure*}[ht]
\centering
\includegraphics[width=0.98\textwidth]{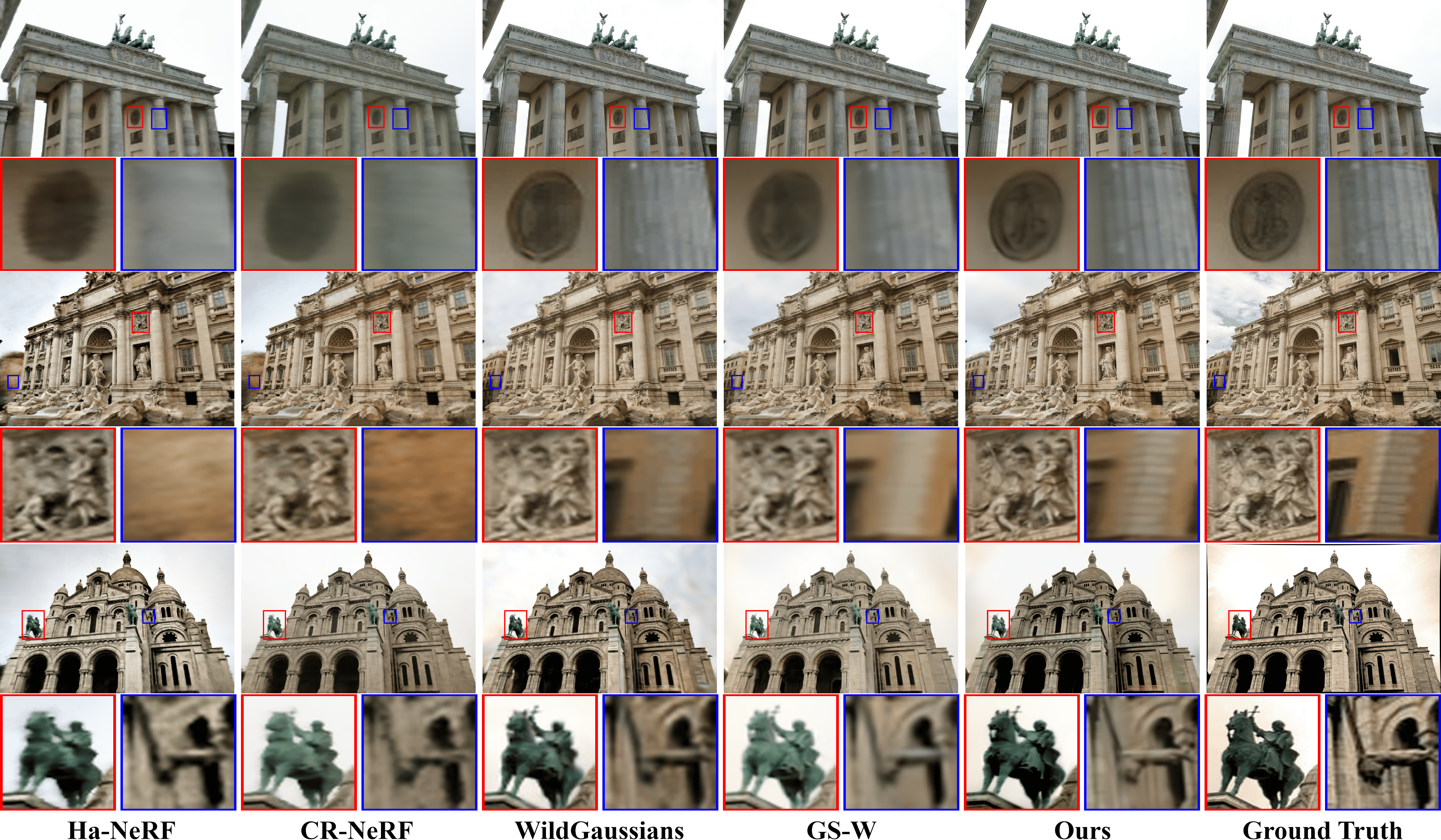}
\caption{Qualitative comparison on three classical unconstrained datasets. Red and blue crops emphasize that SMW-GS can recover finer details.}
\label{fig:qualitative-classical}
\end{figure*}

\begin{table}[t]
\centering
\setlength{\tabcolsep}{5pt}
\caption{Rendering speed comparison on three datasets with a resolution of $800 \times 800$ using a single RTX 3090 GPU, measuring performance in FPS. \textbf{Bold} and \underline{underlined} values correspond to the best and the second-best value, respectively.}
\label{tab:speed}
\begin{tabular}{ccccc}
\toprule
    \multirow{3}{*}{Method} & \multicolumn{3}{c}{Render Speed (FPS)} & \multirow{3}{*}{\makecell[c]{Reconstruction\\Time (Hour)}}\\ 
    \cmidrule(lr){2-4}
    & \textit{\makecell[c]{Brandenburg\\Gate}} & \textit{\makecell[c]{Sacre\\Coeur}} & \textit{\makecell[c]{Trevi\\Fountain}} & \\
\midrule
    Ha-NeRF & 0.0489 & 0.0497 & 0.0498 & 71.6\\
    CR-NeRF & 0.0445 & 0.0447 & 0.0446 & 101\\
    WildGaussians & 36.10 & 40.42 & 18.54 & 6.91\\
    GS-W & \underline{54.49} & \underline{60.31} & \underline{39.99} & \underline{2.83}\\
    Ours & \textbf{64.43} & \textbf{92.08} & \textbf{61.01} & \textbf{2.60}\\
\bottomrule
\end{tabular}
\end{table}

\begin{table*}[ht]
\setlength{\tabcolsep}{5pt}
\centering
\caption{Quantitative results on four real large-scale unconstrained datasets. \textbf{Bold} and \underline{underlined} values correspond to the best and the second-best value, respectively. Our method outperforms the previous methods across all datasets on PSNR , SSIM, and LPIPS.}
\label{tab:quantitative-real-large-scale}
\begin{tabular}{cccccccccccccccc}
\toprule
\multirow{2}{*}{Method} & \multicolumn{3}{c}{\textit{Rubble}} & \multicolumn{3}{c}{\textit{Building}} & \multicolumn{3}{c}{\textit{Residence}} & \multicolumn{3}{c}{\textit{Sci-Art}}\\ 
\cmidrule(lr){2-4} \cmidrule(lr){5-7} \cmidrule(lr){8-10} \cmidrule(lr){11-13} 
& PSNR $\uparrow$ & SSIM $\uparrow$ & LPIPS $\downarrow$ & PSNR $\uparrow$ & SSIM $\uparrow$ & LPIPS $\downarrow$ & PSNR $\uparrow$ & SSIM $\uparrow$ & LPIPS $\downarrow$ & PSNR $\uparrow$ & SSIM $\uparrow$ & LPIPS $\downarrow$ \\
\midrule
WildGaussian & 25.64 & 0.724 & 0.274 & 21.75 & 0.688 & 0.288 & 22.98 & 0.768 & 0.214 & 22.85 & 0.760 & 0.221 \\
GS-W & 24.93 & 0.757 & 0.233 & 23.68 & 0.786 & 0.187 & \underline{24.15} & 0.807 & 0.179 & \underline{25.76} & 0.843 & 0.209 \\
\midrule
VastGaussian & 26.08 & 0.827 & 0.134 & 23.22 & 0.789 & 0.144 & 22.38 & 0.776 & 0.145 & 23.04 & 0.784 & 0.159 \\
CityGaussian & 25.59 & 0.810 & 0.146 & 21.96 & 0.784 & 0.160 & 21.62 & 0.817 & 0.172 & 21.54 & 0.781 & 0.176 \\
Momentum-GS & \underline{26.87} & \underline{0.836} & \underline{0.117} & \underline{24.14} & \underline{0.817} & \underline{0.126} & 23.18 & \underline{0.832} & \underline{0.117} & 24.06 & \underline{0.854} & \underline{0.147} \\
\midrule
Ours & \textbf{29.03} & \textbf{0.877} & \textbf{0.098} & \textbf{25.36} & \textbf{0.822} & \textbf{0.119} & \textbf{26.59} & \textbf{0.867} & \textbf{0.091} & \textbf{27.61} & \textbf{0.859} & \textbf{0.147} \\
\bottomrule
\end{tabular}
\end{table*}

\begin{figure*}[ht]
\centering
\includegraphics[width=0.98\textwidth]{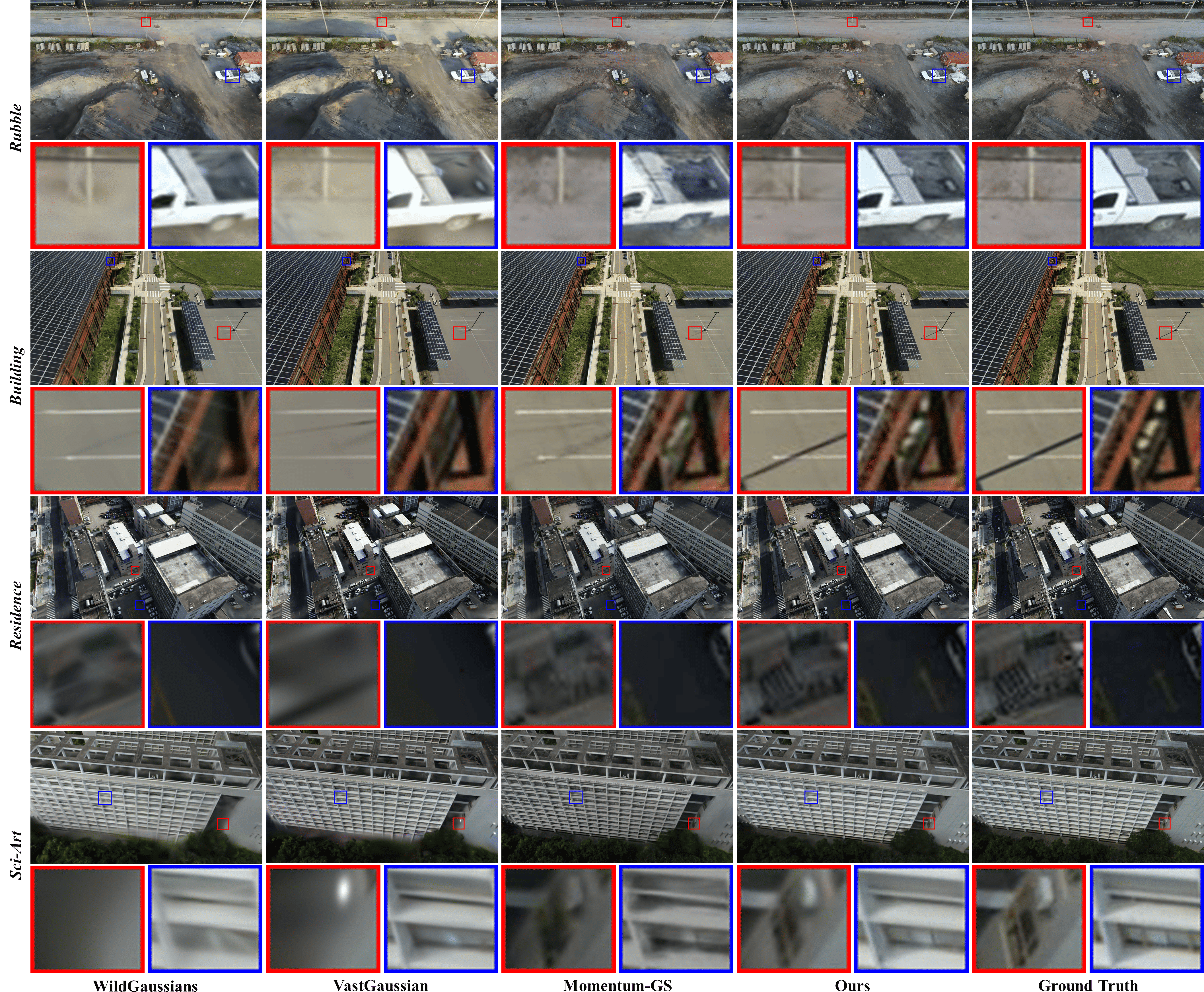}
\caption{Qualitative comparison on four real large-scale unconstrained datasets. Red and blue crops emphasize that SMW-GS can recover finer details.}
\label{fig:qualitative-large-scale}
\end{figure*}

\begin{figure*}[ht]
\centering
\includegraphics[width=0.98\textwidth]{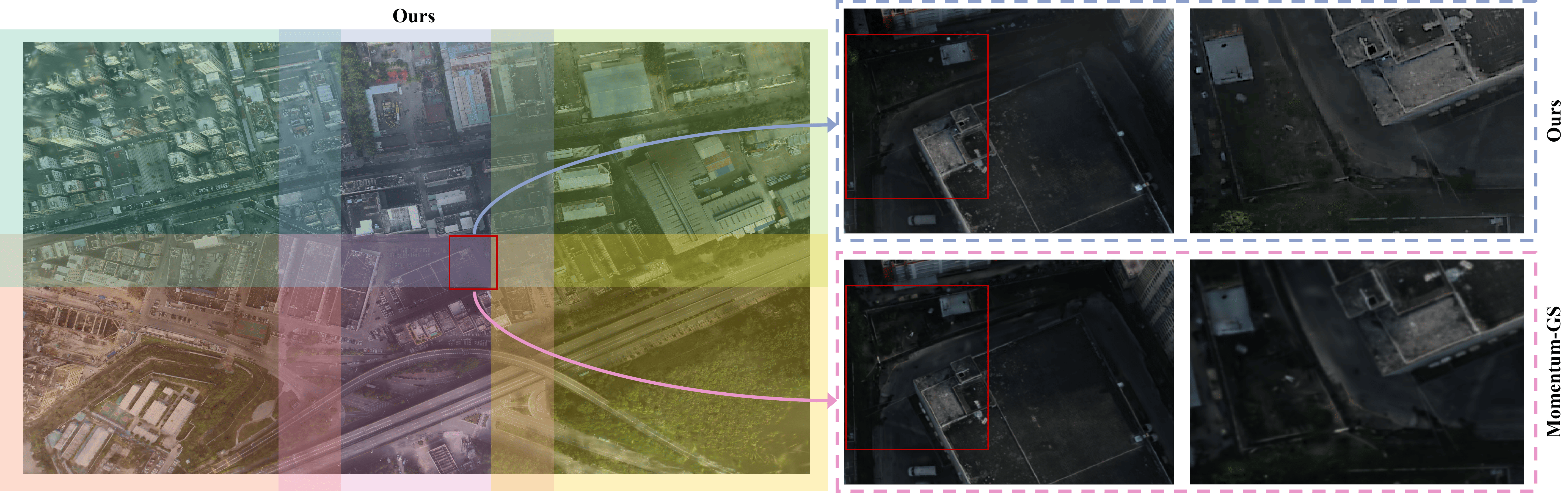}
\caption{Qualitative results on the \textit{Residence} scene at block boundaries. Our method significantly enhances the visual quality at block boundaries, preserving sharpness and structural consistency. The top row on the right displays results from our method, while the bottom row illustrates results from Momentum-GS.}
\label{fig:coarse-to-fine}
\end{figure*}

\subsection{Implementation Details}
This section provides a detailed overview of the hyperparameter configurations and network settings used in SMW-GS, along with information about the baselines used for comparison.

We develop our method based on the original implementation of Scaffold-GS. In our setup: Gaussians per voxel \( k =10\), frustum samples \( k_s = 1 \), wavelet dim \( M = 1 \), intrinsic feature dim \( n_v = 48 \), refined feature dim \( n_r = 32 \), global feature dim \( n_g = 16 \). Learning rates for \(nc_i\) and \(bc_i\) decay from \(1\times 10^{-4}\) to \(1\times 10^{-5}\). Optimization uses Adam with \(\lambda_{\text{SSIM}} = 0.2\), \(\lambda_{1} = 0.8\), \(\lambda_{proj} = 0.01\), and \(\lambda_{vol} = 0.01\). Other hyper-parameters are set according to the guidelines of Scaffold-GS.

We use a ResNet-18 encoder (up to the layer before AdaptiveAvgPool2d), with frozen batch normalization.  The global feature MLP \(MLP^G\) has one hidden layer of size \( 2n_g \), and it ultimately outputs the global appearance feature. The UNet decoder has four upsampling blocks with residual connections, followed by a final convolutional layer projecting to \( n_r \). These modules are trained with a learning rate decaying from \(1\times 10^{-4}\) to \(1\times 10^{-6}\). Since each Gaussian is trained per image, only ReLU activations are used (no batch norm) in the decoder.

The proposed Hierarchical Residual Fusion Network (HRFN) consists of four MLPs, denoted as \( \mathcal{M}^H = \{ \mathcal{M}^H_1, \mathcal{M}^H_2, \mathcal{M}^H_3, \mathcal{M}^H_4 \} \), with the number of hidden units for each set as follows: \{128, 96\}, \{96, 64\}, \{48, 48\}, and \{48\}. Learning rate decays from \(5\times 10^{-4}\) to \(5\times 10^{-5}\).
 
For a fair comparison, VastGaussian incorporates the Decoupled Appearance Modeling module. VastGaussian, Momentum-GS and WildGaussians optimize the appearance embeddings on test images while keeping other parameters frozen. CityGaussian uses the official implementation.

\subsection{Comparison Results}
\subsubsection{Classical Unconstrained Data Evaluation}
The quantitative results for the three classical unconstrained scenes, presented in Tab. \ref{tab:quantitative-classical}, highlight the effectiveness of the proposed SMW-GS method.
Ha-NeRF and CR-NeRF show improvements over earlier baselines but remain limited in capturing local contextual cues essential for modeling diverse scene points.
Similarly, WildGaussians struggle with appearance modeling due to their reliance on global appearance embeddings.
GS-W mitigates these limitations through adaptive sampling of local features, enabling a more precise representation of fine-grained details. This targeted approach is reflected in its consistently superior performance across evaluation metrics.
The proposed SMW-GS method advances further by seamlessly integrating long-range contextual information with detailed local features within Gaussian representations. By enhancing multi-scale information fusion and effectively capturing high-frequency details, SMW-GS achieves notable improvements. These advancements are evident in significant gains in PSNR, surpassing GS-W by 1.41 dB, 1.40 dB, and 1.16 dB across the three evaluated scenes, underscoring its robust performance in handling complex unconstrained scenarios.

The qualitative results in Fig. \ref{fig:qualitative-classical} vividly illustrate the advantages of our approach. For instance, our method captures finer details and more accurate colors in the reliefs of the Trevi Fountain and the bronze statues at Sacre Coeur, surpassing the capabilities of existing techniques. While current methods often struggle with accurately representing intricate scene details and complex textures, our approach excels by leveraging micro-macro wavelet-based sampling to enhance feature extraction. This technique effectively integrates frequency-domain and multi-scale information, while the hierarchical fusion of structured features facilitates the precise recovery of appearance details and clear structural representation.

To assess the training efficiency and rendering performance during inference, we conducted experiments on three datasets with an image resolution of $800 \times 800$, using a single RTX 3090 GPU to compute the average rendering time per image. The overall inference time includes the feature extraction time for reference images in Ha-NeRF, CR-NeRF, GS-W, and our method. As summarized in Tab. \ref{tab:speed}, our approach not only ensures fine-grained modeling of image appearance but also demonstrates excellent rendering speed, being nearly 1.5 times faster than existing Gaussian-based methods. Furthermore, we evaluated the reconstruction time (in hours) required for training, with results showing that our method enables efficient training while maintaining high-quality outcomes.

\begin{table}[t]
\centering
\setlength{\tabcolsep}{2pt}
\caption{Storage usage (in GB) across four real large-scale scenes and rendering speed (in FPS) for four datasets at a resolution of $1,920 \times 1,080$, measured on a single RTX 3090 GPU.}
\label{tab:storage}
\begin{tabular}{ccccccccc}
\toprule
\multirow{2}{*}{Method} & \multicolumn{2}{c}{\textit{Rubble}} &  \multicolumn{2}{c}{\textit{Building}} &  \multicolumn{2}{c}{\textit{Residence}} &  \multicolumn{2}{c}{\textit{Sci-Art}} \\
\cmidrule(lr){2-3} \cmidrule(lr){4-5} \cmidrule(lr){6-7} \cmidrule(lr){8-9}
& Storage & FPS & Storage & FPS & Storage & FPS & Storage & FPS \\
\midrule
CityGaussian& 2.22 & \textbf{50.48} & 3.07 & \underline{32.99} & 2.49 & \textbf{40.12} & \textbf{0.90} & \textbf{43.32} \\
Momentum-GS & \underline{1.63}  & 37.87 & \underline{2.36} & 29.58 & \underline{1.89} & 25.75 & 0.96 & 23.15 \\
Ours & \textbf{1.35} & \underline{41.26} & \textbf{1.86} & \textbf{34.90} & \textbf{1.41} & \underline{37.45} & \underline{0.91}  & \underline{30.24} \\
\bottomrule
\end{tabular}
\end{table}

\begin{table*}[ht]
\setlength{\tabcolsep}{5pt}
\centering
\caption{Quantitative results on synthetic large-scale unconstrained datasets. \textbf{Bold} and \underline{underlined} values correspond to the best and the second-best value, respectively. Our method outperforms the previous methods across all datasets on PSNR , SSIM, and LPIPS.}
\label{tab:quantitative-synthetic-large-scale}
\begin{tabular}{cccccccccccccccc}
\toprule
\multirow{2}{*}{Method} & \multicolumn{3}{c}{\textit{Block\_A$\ast$}} & \multicolumn{3}{c}{\textit{Block\_E$\ast$}} & \multicolumn{3}{c}{\textit{Block\_A}} & \multicolumn{3}{c}{\textit{Block\_E}}\\ 
\cmidrule(lr){2-4} \cmidrule(lr){5-7} \cmidrule(lr){8-10} \cmidrule(lr){11-13} 
& PSNR $\uparrow$ & SSIM $\uparrow$ & LPIPS $\downarrow$ & PSNR $\uparrow$ & SSIM $\uparrow$ & LPIPS $\downarrow$ & PSNR $\uparrow$ & SSIM $\uparrow$ & LPIPS $\downarrow$ & PSNR $\uparrow$ & SSIM $\uparrow$ & LPIPS $\downarrow$ \\
\midrule
WildGaussians & 22.28 & 0.665 & 0.436 & 21.81 & 0.627 & 0.459 & 26.04 & 0.737 & 0.328 & 24.48 & 0.718 & 0.385 \\
GS-W & \underline{23.24} & 0.648 & 0.458 & \underline{22.29} & 0.603 & 0.564 & 25.77 & 0.729 & 0.344 & 25.33 & 0.734 & 0.309 \\
\midrule
VastGaussian & 21.72 & 0.713 & 0.229 & 20.45 & 0.689 & 0.189 & 27.57 & 0.845 & 0.160 & 27.30 & 0.861 & 0.129 \\
CityGaussian & 21.64 & 0.765 & 0.256 & 21.84 & 0.769 & 0.2 & 27.19 & 0.815 & 0.202 & 27.07 & 0.829 & 0.145 \\
Momentum-GS & 21.55 & \underline{0.776} & \underline{0.181} & 20.85 & \underline{0.786} & \underline{0.156} & \underline{28.13} & \underline{0.852} & \underline{0.111} & \underline{27.74} & \underline{0.867} & \underline{0.092} \\
\midrule
Ours & \textbf{28.06} & \textbf{0.849} & \textbf{0.108} & \textbf{27.21} & \textbf{0.846} & \textbf{0.104} & \textbf{28.73} & \textbf{0.859} & \textbf{0.097} & \textbf{28.48} & \textbf{0.877} & \textbf{0.086} \\

\bottomrule
\end{tabular}
\end{table*}

\subsubsection{Real Large-scale Unconstrained Data Evaluation}
Table \ref{tab:quantitative-real-large-scale} presents the quantitative results on four real-world, large-scale, and unconstrained datasets, underscoring the scalability and effectiveness of the proposed SMW-GS method in handling expansive scenes. While WildGaussians and GS-W exhibit commendable performance in classic in-the-wild scenarios, they face challenges in generalizing to large-scale environments. Despite leveraging decoupled appearance modeling and achieving relatively high PSNR on UrbanScene3D, their overall reconstruction quality falls short, particularly in structure-aware metrics like SSIM.
Among methods specifically designed for large-scale scenes, VastGaussian incorporates appearance embeddings and a CNN-based appearance adjustment module but struggles to establish accurate image-level mappings between rendered and real images. Similarly, Momentum-GS, which employs a simple appearance embedding, encounters difficulties with effective appearance disentanglement in expansive scenes.
In contrast, our SMW-GS method achieves significant improvements by seamlessly integrating long-range contextual information with fine-grained local features within the Gaussian representation. Additionally, the scale-up strategy enhances the representation of complex scenes, enabling superior reconstruction performance. Across the four datasets, SMW-GS consistently outperforms existing in-the-wild and large-scale reconstruction methods, surpassing the previous best, Momentum-GS, by 2.16 dB, 1.22 dB, 3.41 dB, and 3.55 dB in PSNR, respectively.

The qualitative results in Fig. \ref{fig:qualitative-large-scale} highlight the distinct advantages of our method in large-scale scene reconstruction. By effectively integrating frequency-domain and multi-scale information and ensuring sufficient Gaussian-level supervision during large-scale training, our approach consistently outperforms existing methods. The reconstructed scenes demonstrate superior visual fidelity and highly accurate geometric details.
For example, our method precisely captures subtle features, such as the shadow of a lamppost in the \textit{Building} scene, and accurately reconstructs intricate structures like the staircases in the \textit{Residence} scene, delivering significantly better performance compared to prior techniques. 

Additionally, Fig. \ref{fig:coarse-to-fine} illustrates the robustness of our approach in handling block boundary transitions within the \textit{Residence} scene. The top row on the right presents results from our method, whereas the bottom row shows those from Momentum-GS. While Momentum-GS suffers from blurred reconstructions at region edges, our method ensures sharpness and structural consistency. This improvement can be attributed to our Point-Statistics-Guided Camera Partitioning strategy, which provides enhanced supervision for Gaussians situated near block boundaries.

Furthermore, we report the corresponding storage usage in Tab. \ref{tab:storage}, where our method achieves superior reconstruction quality with reduced storage consumption. This efficiency stems from compressing appearance information into the appearance disentanglement network, significantly reducing storage requirements while maintaining high reconstruction accuracy.

\begin{figure*}[!ht]
\centering
\includegraphics[width=0.98\textwidth]{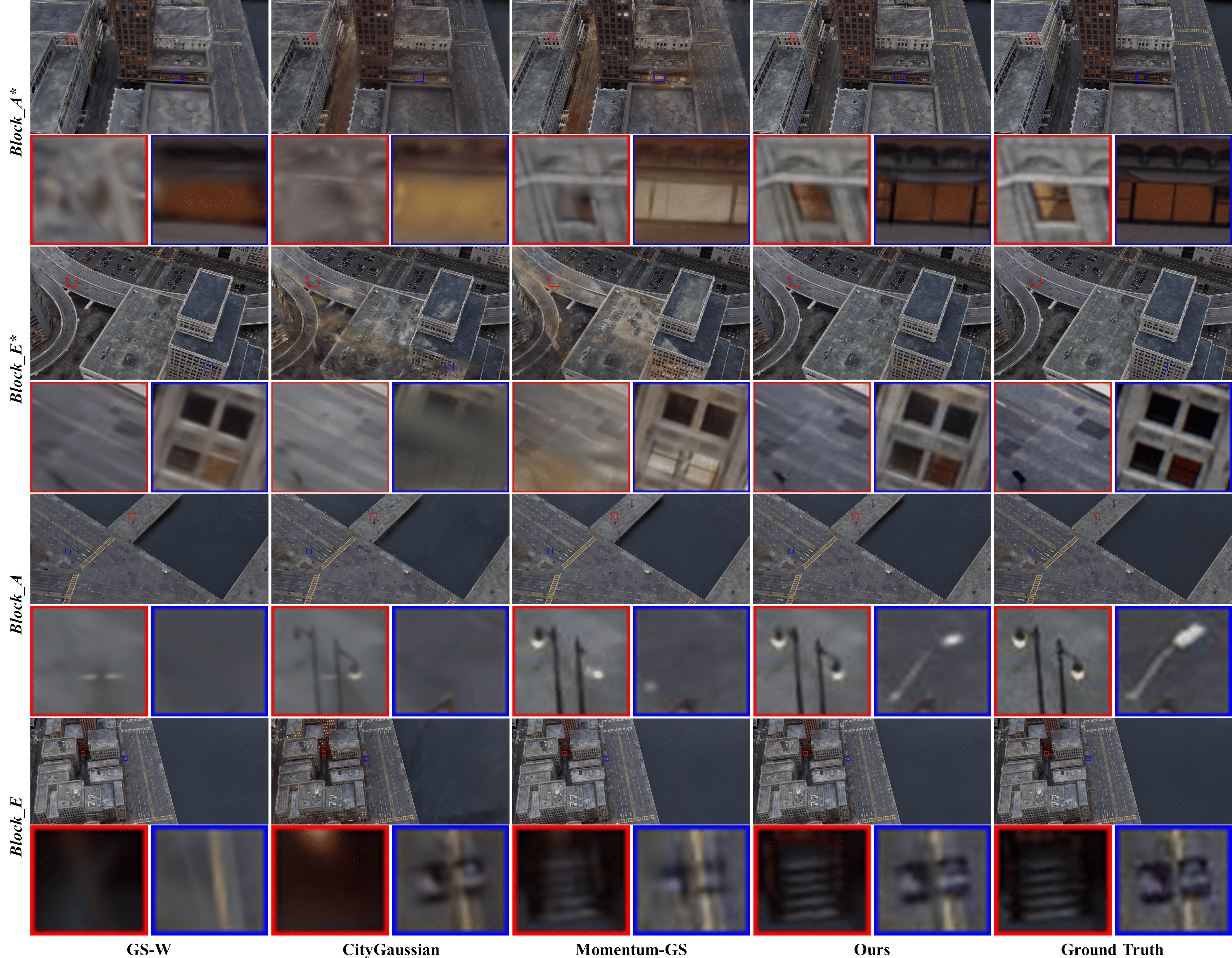}
\caption{Qualitative comparison on synthetic large-scale unconstrained datasets. Red and blue crops highlight that SMW-GS effectively disentangles complex appearance, resulting in more accurate color reproduction and finer detail restoration.}
\label{fig:qualitative-syn-large-scale}
\end{figure*}

\begin{figure*}[ht]
\centering
\includegraphics[width=0.98\textwidth]{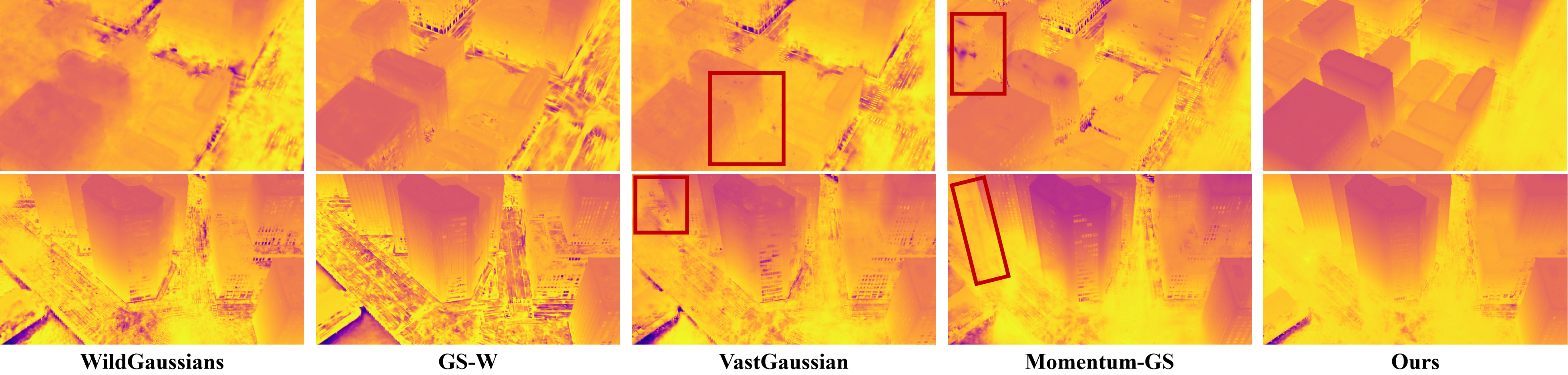}
\caption{Qualitative comparison of depth maps generated by different methods, displaying rendering viewpoints interpolated between training views with significant appearance differences. Red insets highlight artifacts and blurry geometric reconstructions in competing methods.}
\label{fig:matrixcity_depth}
\end{figure*}

\subsubsection{Synthetic Large-scale Data Evaluation}
Tab. \ref{tab:quantitative-synthetic-large-scale} presents the quantitative results for four synthetic large-scale scenes, including two with significant appearance variations \textit{Block\_A$\ast$} and \textit{Block\_E$\ast$} and two with consistent appearance (\textit{Block\_A} and \textit{Block\_E}). These findings highlight the scalability and effectiveness of our proposed SMW-GS method in handling large-scale environments. Our approach demonstrates robust performance in consistent-appearance scenarios while effectively managing significant appearance variation through advanced disentanglement techniques.

As previously observed, methods like WildGaussians and GS-W struggle with generalization in large-scale settings. Although these methods achieve relatively higher PSNR on \textit{Block\_A$\ast$} and \textit{Block\_E$\ast$} due to their appearance modeling, their reconstruction quality suffers a significant decline, particularly in structure-aware metrics. Additionally, the performance gap between scenes with consistent appearance (\textit{Block\_A} and \textit{Block\_E}) and those with appearance variation (\textit{Block\_A$\ast$} and \textit{Block\_E$\ast$}) underscores their inability to scale effectively.

Methods designed specifically for large-scale scenarios, such as VastGaussian and Momentum-GS, face even steeper declines in reconstruction quality when transitioning from consistent to varied appearance settings. These results emphasize the challenges of adapting in-the-wild methods to large-scale environments and the importance of scalable, robust solutions.

In contrast, our SMW-GS method excels in both scalability and robustness. It maintains high performance across consistent and varied appearance scenarios, with minimal performance drop between \textit{Block\_A} and \textit{Block\_A$\ast$} or \textit{Block\_E} and \textit{Block\_E$\ast$}. This demonstrates the superior disentanglement and adaptability of our framework. Across all four synthetic datasets, SMW-GS surpasses Momentum-GS in PSNR by 6.51 dB, 6.36 dB, 0.60 dB, and 0.74 dB, respectively, significantly outperforming both in-the-wild and large-scale baseline methods.

The qualitative results presented in Fig. \ref{fig:qualitative-syn-large-scale} further highlight the superiority of our method in reconstructing large-scale scenes, both under significant appearance variations and in appearance-consistent environments.
In challenging scenes such as \textit{Block\_A$\ast$} and \textit{Block\_E$\ast$}, CityGaussian struggles to effectively disentangle appearance components in complex conditions, leading to pronounced visual artifacts. Similarly, Momentum-GS, relying solely on globally learnable appearance embeddings, fails to handle intricate appearance variations, resulting in noticeable color inconsistencies and artifacts in the rendered images. While GS-W demonstrates relatively consistent appearance matching with the ground truth, its limited scalability to large scenes and inability to reconstruct fine details are evident.
In contrast, our method delivers superior fidelity and precision in both appearance and geometric detail across all four scenes, significantly surpassing existing methods. For instance, our approach captures the intricate staircase structures at the base of buildings in the \textit{Block\_E} scene and faithfully reproduces fine details such as windows and pavement in \textit{Block\_A$\ast$} and \textit{Block\_E$\ast$}, achieving appearance nearly indistinguishable from the ground truth and demonstrating a clear advantage over prior techniques.

Additionally, we provide a qualitative comparison of depth maps across different methods, with rendering viewpoints interpolated between training views exhibiting substantial appearance variations. WildGaussians and GS-W, which are not optimized for large-scale unconstrained scenes, produce depth maps plagued by significant blurring and noise, resulting in irregular and low-quality depth reconstructions. Momentum-GS and VastGaussian, despite attempting to address large-scale reconstruction, employ simplistic strategies that fail to manage illumination variation effectively, leading to severe artifacts and blurred results, as highlighted by the red insets. In contrast, our method employs explicit disentanglement of appearance into three structured feature components, enabling robust and consistent geometric reconstruction across diverse appearance conditions, achieving state-of-the-art geometric fidelity.

\begin{table*}[ht]
\setlength{\tabcolsep}{5pt}
\centering
\caption{Ablation studies on decoupling module. \textbf{Bold} and \underline{underlined} values correspond to the best and the second-best value.}
\begin{tabular}{ccccccccccccc}
\toprule
\multirow{2}{*}{Method} & \multicolumn{3}{c}{\textit{Brandenburg Gate}} & \multicolumn{3}{c}{\textit{Trevi Fountain}} & \multicolumn{3}{c}{\textit{Rubble}} & \multicolumn{3}{c}{\textit{Block\_A$\ast$}} \\

\cmidrule(lr){2-4} \cmidrule(lr){5-7} \cmidrule(lr){8-10} \cmidrule(lr){11-13} 
& PSNR $\uparrow$ & SSIM $\uparrow$ & LPIPS $\downarrow$ & PSNR $\uparrow$ & SSIM $\uparrow$ & LPIPS $\downarrow$ & PSNR $\uparrow$ & SSIM $\uparrow$ & LPIPS $\downarrow$ & PSNR $\uparrow$ & SSIM $\uparrow$ & LPIPS $\downarrow$\\
\midrule
only micro & 28.49 & 0.939 & 0.059 & 23.47  & 0.809 & 0.129 & \underline{28.72} & \underline{0.871} & 0.105 & \underline{27.75} & \underline{0.847} & \underline{0.112} \\
w/o WS & \underline{28.88}  & \underline{0.939}  & \underline{0.055} & 23.38  & 0.810 & 0.127 & 28.34 & 0.857 & 0.109 & 27.59 & 0.841 & 0.115\\
w/o HRFN & 28.69  & 0.936  & 0.057 & \underline{23.68} & \underline{0.813} & \underline{0.125} & 28.36 & 0.864 & \underline{0.104} & 27.67 & 0.846 & 0.114\\
\midrule
Full model &  \textbf{29.37} &  \textbf{0.942} & \textbf{0.052} & \textbf{24.07} &  \textbf{0.821} &  \textbf{0.120} & \textbf{29.03} & \textbf{0.877} & \textbf{0.098} & \textbf{28.06} & \textbf{0.849} & \textbf{0.108} \\
\bottomrule
\end{tabular}
\label{tab:ablation}
\end{table*}

\begin{figure}[t]    
\centering
\includegraphics[width=0.98\columnwidth]{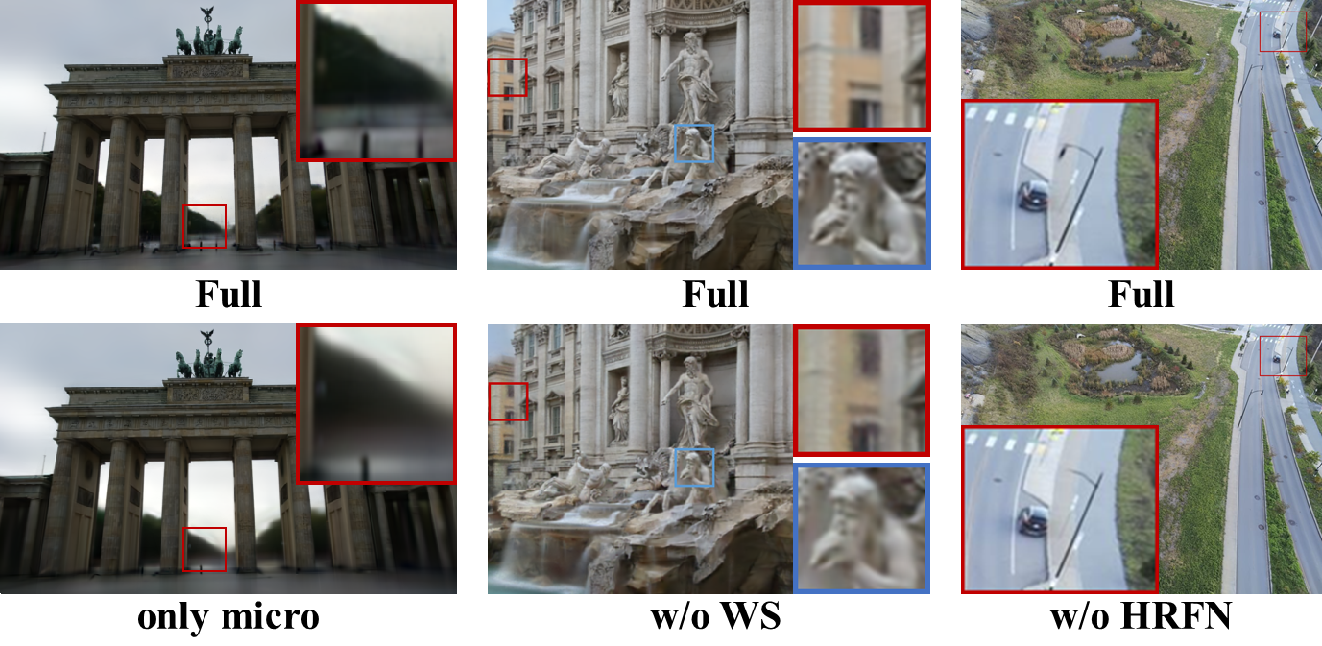}
\caption{Ablation studies by visualization. The images demonstrate the effects of key components, including Micro-macro Projection (MP), Wavelet-based Sampling (WS), and the Hierarchical Residual Feature Network (HRFN), on reconstruction quality and detail retention.}

\label{fig:ablation}
\end{figure}

\subsection{Component Analysis}
\subsubsection{Ablation Study on Appearance Decoupling Module}

\begin{figure}[t]    
  \centering        
  \subfloat[Sampling attention.]
  {
      \includegraphics[width=0.33\columnwidth]{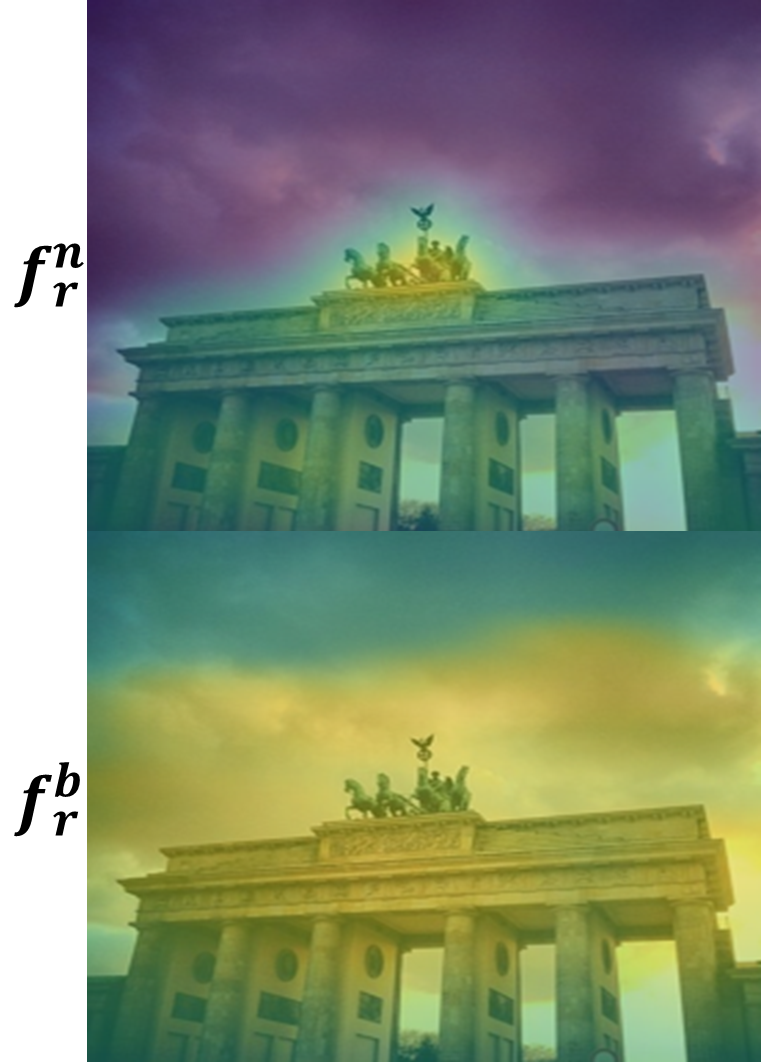}
      \label{fig:feat_attention}
  }
  \subfloat[Sampling features rendering result.]
  {
      \includegraphics[width=0.60\columnwidth]{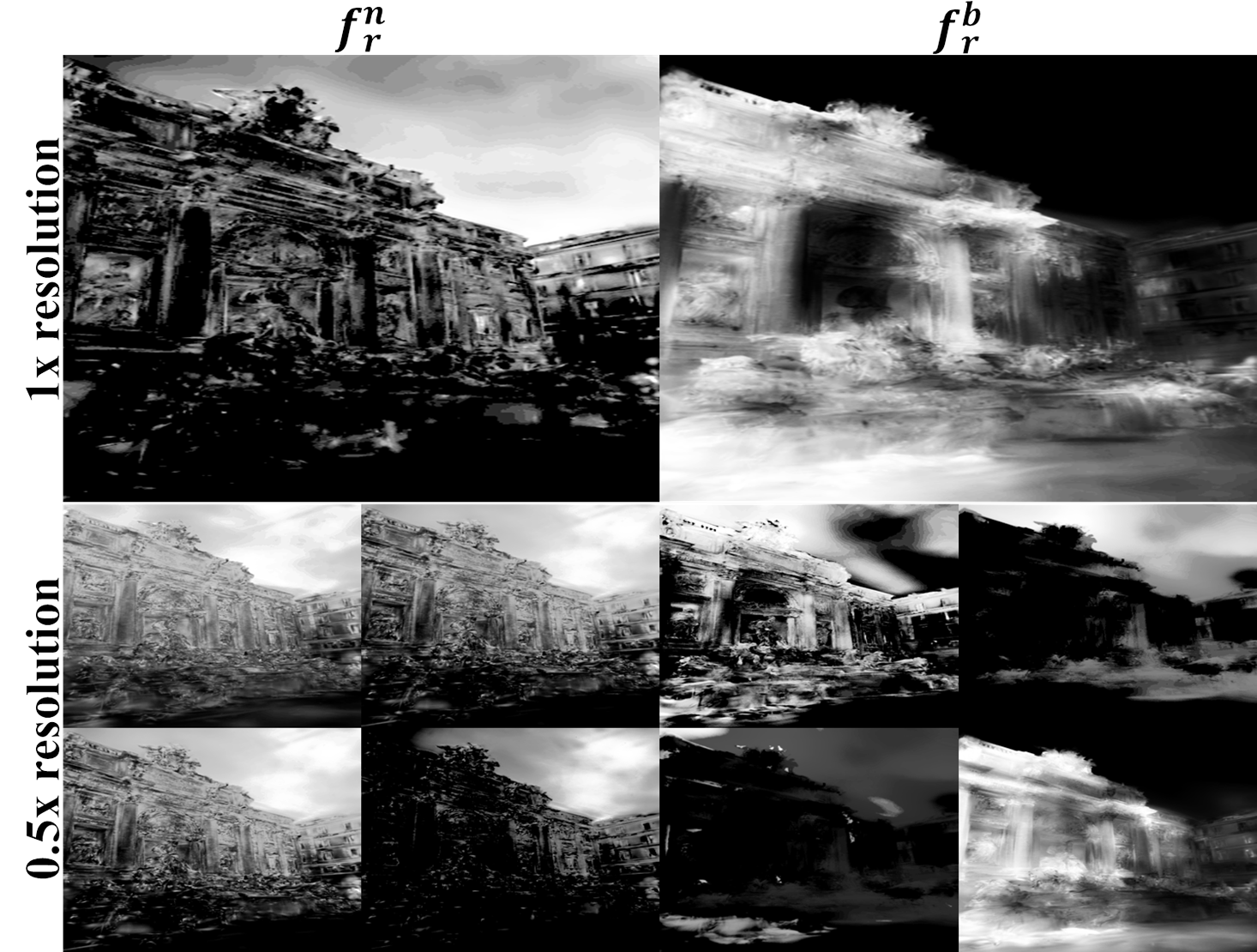}
      \label{fig:feat_sampling}
  }
  \caption{Visualization of sampling analysis. (a) The attention maps generated by projecting sampling positions onto corresponding camera images. (b) The refined features \( f^n_r \) and \( f^b_r \) highlight the ability to integrate high-resolution textures and low-texture regions across multiple frequency bands.}
 
  \label{fig:sampling-analysis}  
\end{figure}

The ablation study results conducted on classical unconstrained datasets, real large-scale datasets, and synthetic large-scale datasets are summarized in Tab. \ref{tab:ablation}. Key findings are as follows:

\textit{i.} Micro-macro Projection (MP):
MP significantly enhances the diversity of refined appearance sampling, allowing Gaussians to more accurately capture appearance features and local contextual information. As illustrated in Fig. \ref{fig:ablation}, relying solely on micro projection (without full MP) results in noticeable blurring of distant objects and geometric inaccuracies. For example, the cylindrical pillar on the right side of the ground is poorly reconstructed when MP is omitted.
\textit{ii.} Wavelet-based Sampling (WS):
WS further refines attention to high-frequency and multi-scale information, resulting in superior reconstruction and rendering quality. When WS is excluded, there is a marked loss of detail, evidenced by the blurring of the Trevi Fountain sculptures and a 0.69 dB decrease in PSNR. This effect becomes even more pronounced in large-scale scenes, which are more sensitive to multi-scale information. For instance, the SSIM on the \textit{rubble} dataset decreases by 0.02 without WS.
\textit{iii.} Hierarchical Residual Feature Network (HRFN):
HRFN provides a more effective integration of features across different levels, enabling a comprehensive fusion of diverse information compared to simple concatenation. This results in a 0.68 dB increase in PSNR on the \textit{Brandenburg Gate} dataset and a 0.39 dB increase on the \textit{Block\_A$\ast$} dataset. Furthermore, HRFN improves the accuracy of color predictions, benefiting the reconstruction of fine-grained structures. For example, it enhances the quality of reconstructed street lamps and other intricate details.
The combined impact of these components ensures robust and detailed scene reconstruction across diverse datasets, validating the effectiveness of the proposed appearance decoupling module.

\begin{table*}[ht]
\centering
\caption{Quantitative results for different values of \(M\) across three datasets. \(M\) denotes the number of wavelet downsampling steps. \(M=1\) achieves the best balance between efficiency and effectiveness. \textbf{Bold} indicates the best result.}

\begin{tabular}{cccccccccc}
\toprule
\multirow{2}{*}{Method} & \multicolumn{3}{c}{\textit{Brandenburg Gate}} & \multicolumn{3}{c}{\textit{Sacre Coeur}} & \multicolumn{3}{c}{\textit{Trevi Fountain}} \\ 
\cmidrule(lr){2-4} \cmidrule(lr){5-7} \cmidrule(lr){8-10} 
& PSNR $\uparrow$ & SSIM $\uparrow$ & LPIPS $\downarrow$ & PSNR $\uparrow$ & SSIM $\uparrow$ & LPIPS $\downarrow$ & PSNR $\uparrow$ & SSIM $\uparrow$ & LPIPS $\downarrow$ \\
\midrule
$M=0$ & 28.88  & 0.939  & 0.055 & 24.35  & 0.895 & 0.075 & 23.38  & 0.810 & 0.127 \\
$M=1$ & \textbf{29.37} & \textbf{0.942} & \textbf{0.052} & \textbf{24.64} & \textbf{0.897} & \textbf{0.073} & \textbf{24.07} & \textbf{0.821} & \textbf{0.120} \\
$M=2$ & 28.94  & 0.938  & 0.054 & 24.41  & 0.897  & 0.076 & 23.40 &  0.810   & 0.127 \\
\bottomrule
\end{tabular}
\label{tab:feat_sampling_all}
\end{table*}

\begin{table*}[ht]
\centering
\caption{Quantitative results for different $k_s$ values across three datasets. \( k_s \) denotes the number of samples per conical frustum cross-section. \( k_s=1 \) balances efficiency and effectiveness. \textbf{Bold} indicates the best result.}

\begin{tabular}{cccccccccc}
\toprule
\multirow{2}{*}{Method} & \multicolumn{3}{c}{\textit{Brandenburg Gate}} & \multicolumn{3}{c}{\textit{Sacre Coeur}} & \multicolumn{3}{c}{\textit{Trevi Fountain}} \\ 
\cmidrule(lr){2-4} \cmidrule(lr){5-7} \cmidrule(lr){8-10} 
& PSNR $\uparrow$ & SSIM $\uparrow$ & LPIPS $\downarrow$ & PSNR $\uparrow$ & SSIM $\uparrow$ & LPIPS $\downarrow$ & PSNR $\uparrow$ & SSIM $\uparrow$ & LPIPS $\downarrow$ \\
\midrule
$k_s=1$ & \textbf{29.37} & \textbf{0.942} & \textbf{0.052} & \textbf{24.64} & \textbf{0.897} & \textbf{0.073} & \textbf{24.07}  & \textbf{0.821}  & \textbf{0.120} \\
$k_s=2$ & 29.02  & 0.938  & 0.057 & 23.96  & 0.885 &  0.081 & 23.74 &  0.815 &  0.124\\
$k_s=3$ & 29.01  &   0.941  & 0.053 & 24.40 &  0.893  & 0.075 &  23.81 & 0.811  & 0.126\\
\bottomrule
\end{tabular}
\label{tab:ks}
\end{table*}

\subsubsection{Analysis of Sampling} 
To analyze our sampling strategy, we project sampling positions onto camera images to form attention maps, where denser regions indicate higher attention. As shown in Fig. \ref{fig:feat_attention}, our method captures fine local details via narrow frustums and integrates long-range context through broader projections.
We further visualize the features of interest by examining the refined narrow features \( f^n_r \) and broad features \( f^b_r \) across different resolutions, as shown in Fig. \ref{fig:feat_sampling}. The \( f^n_r \) features, focused on high-resolution details, adeptly capture local texture intricacies, while features processed at 0.5$\times$ resolution through DWT attend to varying details across different frequency bands. Conversely, the \( f^b_r \) features primarily target low-texture regions, such as water surfaces or specular highlights, which correspond to long-range features.
The combination of \( f^n_r \) and \( f^b_r \) allows our MW sampling approach to effectively model dynamic appearances by capturing both detailed and distinct information on the 2D feature maps.

\subsubsection{Analysis of Wavelet Dimension}
We study the impact of the wavelet decomposition dimension \( M \), which controls how many times the feature map is downsampled. Experiments on three datasets (Tab. \ref{tab:feat_sampling_all}) show that \( M = 1 \), corresponding to sampling at both 1$\times$ resolution and 0.5$\times$ resolution, yields the best performance. Further downsampling (e.g., 0.25$\times$) offers no gain but increases computation. Therefore, we reasonably set \( M \) to 1.
\subsubsection{Analysis of Sampling Number}
We evaluate the number of samples \( k_s \)per conical frustum cross-section on three scenes (Tab. \ref{tab:ks}). Interestingly, \( k_s = 1 \) achieves the best performance, while higher values degrade results and increase cost. 
The reason could be that additional sampling might cause the features of 3D points along the same ray to converge towards a common mean, reducing diversity. Therefore, we set \( k_s = 1 \).

\begin{table*}[ht]
\centering
\caption{Quantitative results of large-scale scene partitioning analysis experiments. we evaluated several variants: no partitioning, partitioning into 4, 6, and 8 blocks, as well as partitioning into 6 blocks without using Point-Statistics-Guided Camera Partitioning, Block-Sensitivity-Aware Camera Partitioning and Rotational Block Training.}
\label{tab:ablation-large-scale}
\begin{tabular}{cccccccccc}
\toprule
\multirow{2}{*}{Method} & \multicolumn{3}{c}{\textit{Rubble}} & \multicolumn{3}{c}{\textit{Block\_A$\ast$}} & \multicolumn{3}{c}{\textit{Block\_E$\ast$}} \\ 
\cmidrule(lr){2-4} \cmidrule(lr){5-7} \cmidrule(lr){8-10}
& PSNR $\uparrow$ & SSIM $\uparrow$ & LPIPS $\downarrow$ & PSNR $\uparrow$ & SSIM $\uparrow$ & LPIPS $\downarrow$ & PSNR $\uparrow$ & SSIM $\uparrow$ & LPIPS $\downarrow$ \\
\midrule
w/o partition & 28.64 & 0.848 & 0.137 & 26.15 & 0.768 & 0.264 & 26.35 & 0.835 & 0.116 \\
\{2, 2\} & \underline{28.98} & 0.856 & 0.128 & 27.16 & 0.837 & 0.116 & 26.84 & 0.841 & 0.113 \\
\{3, 2\} & \textbf{29.03} & \textbf{0.877} & \textbf{0.098}  & \textbf{28.06} & \textbf{0.849} & \textbf{0.108} & \textbf{27.21} & \underline{0.846} & \textbf{0.104} \\
\{4, 2\} & 28.89 & \underline{0.857} & \underline{0.110} & \underline{27.68} & \underline{0.846} & \underline{0.111} & \underline{27.11} & \textbf{0.848} & \underline{0.109} \\
w/o PSG & 28.57 & 0.847 & 0.137 & 27.46 & 0.839 & 0.121 & 26.94 & 0.828 & 0.117 \\
w/o PSG-S1 & 28.51 & 0.849 & 0.135 & 27.64 & 0.842 & 0.122 & 27.07 & 0.836 & 0.114 \\
w/o PSG-S2 & 28.91 & 0.878	& 0.117 & 27.69 & 0.844 & 0.116 & 27.09 & 0.842 & 0.112 \\
w/o RBT & 26.96 & 0.817 & 0.127 & 27.01 & 0.818 & 0.144 & 25.82 & 0.796 & 0.158 \\
\bottomrule
\end{tabular}
\end{table*}



\begin{figure}[t]
\centering
\includegraphics[width=0.98\columnwidth]{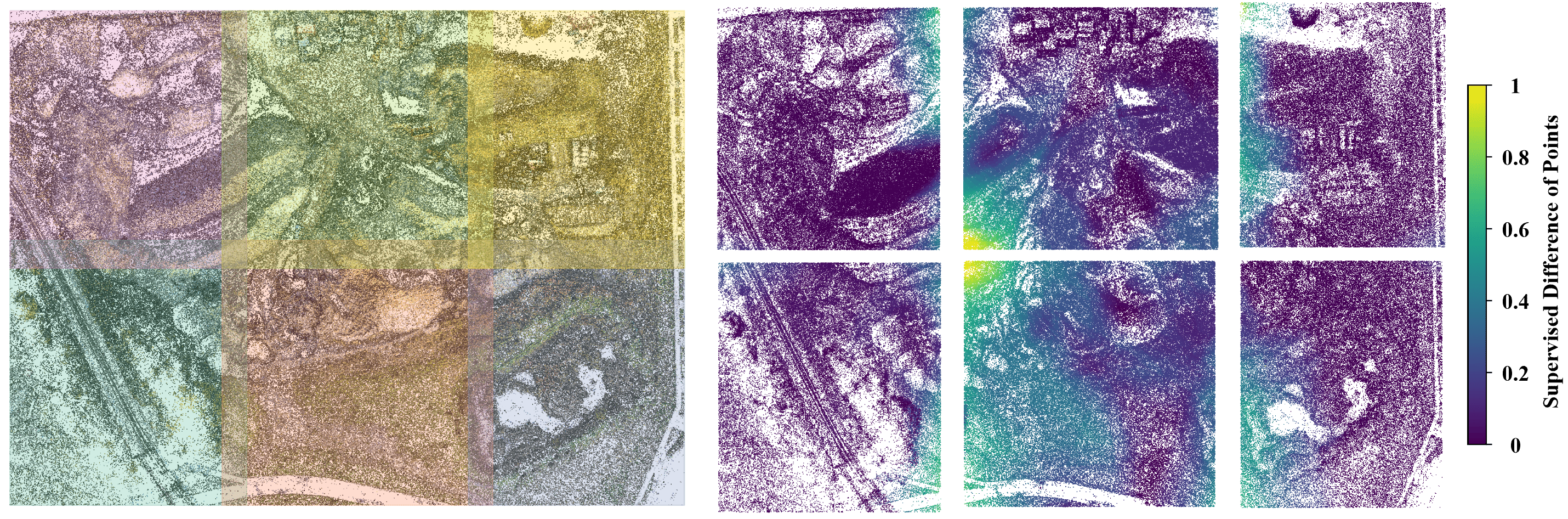}
\caption{Visualization of the \{3,2\} partition strategy applied to the \textit{rubble} scene. \textit{Right:} Normalized per-point increase in supervision counts when using the ``full model" compared to the ``w/o PSG-S1" variant, with values scaled to the range $[0,1]$.}
\label{fig:supervised-difference}
\end{figure}

\subsubsection{Analysis of Large-scale Scene Partitioning}
We conduct experiments on different scene configurations to analyze the contribution of each component in the large-scale scene promotion strategy. The quantitative results are summarized in Tab. \ref{tab:ablation-large-scale}, evaluating the following variants:
``w/o partition" (no spatial partitioning; whole scene trained jointly);
``\{2, 2\}", ``\{3, 2\}", ``\{4, 2\}" (scene partitioned into 2$\times$2, 3$\times$2, and 4$\times$2 blocks, respectively);
``w/o PSG" (3$\times$2 partition, using CityGaussian's partitioning instead of Point-Statistics-Guided (PSG) Camera Partitioning);
``w/o PSG-S1" and ``w/o PSG-S2" (3$\times$2 partition, disabling Stage 1 or Stage 2 of PSG, respectively);
``w/o RBT" (3$\times$2 partition, disabling Rotational Block Training (RBT)).


\begin{figure*}[ht] 
    \centering 
    \includegraphics[width=0.98\textwidth]{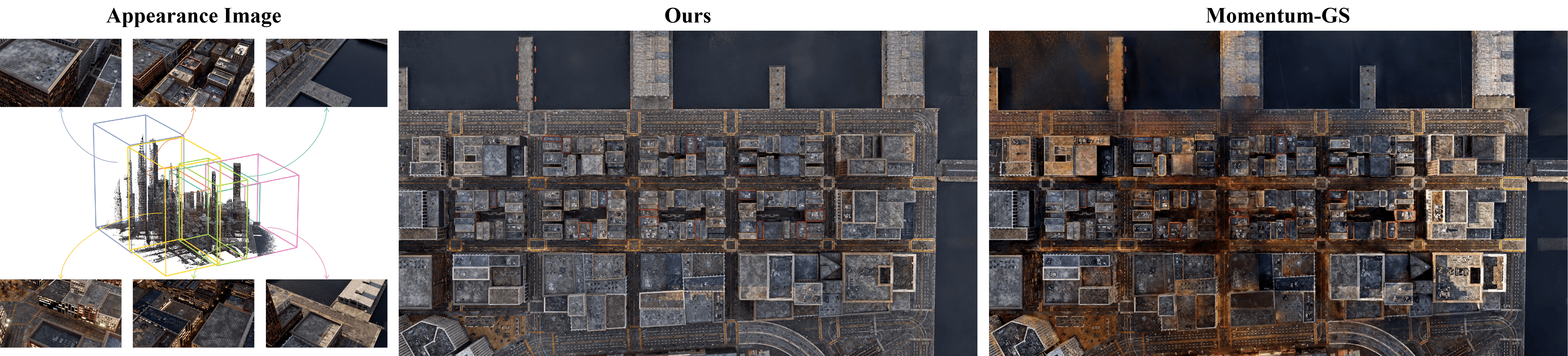}
 \caption{Reconstruction under six blocks with varying appearances. \textit{Left:} Reference images showing distinct appearances used by the six different blocks. \textit{Middle:} Results from our method. \textit{Right:} Results from Momentum-GS. Our method achieves consistent appearance reconstruction across the entire scene.}
    \label{fig:large_appearance}
\end{figure*}

\begin{figure*}[ht] 
    \centering 
    \includegraphics[width=0.98\textwidth]{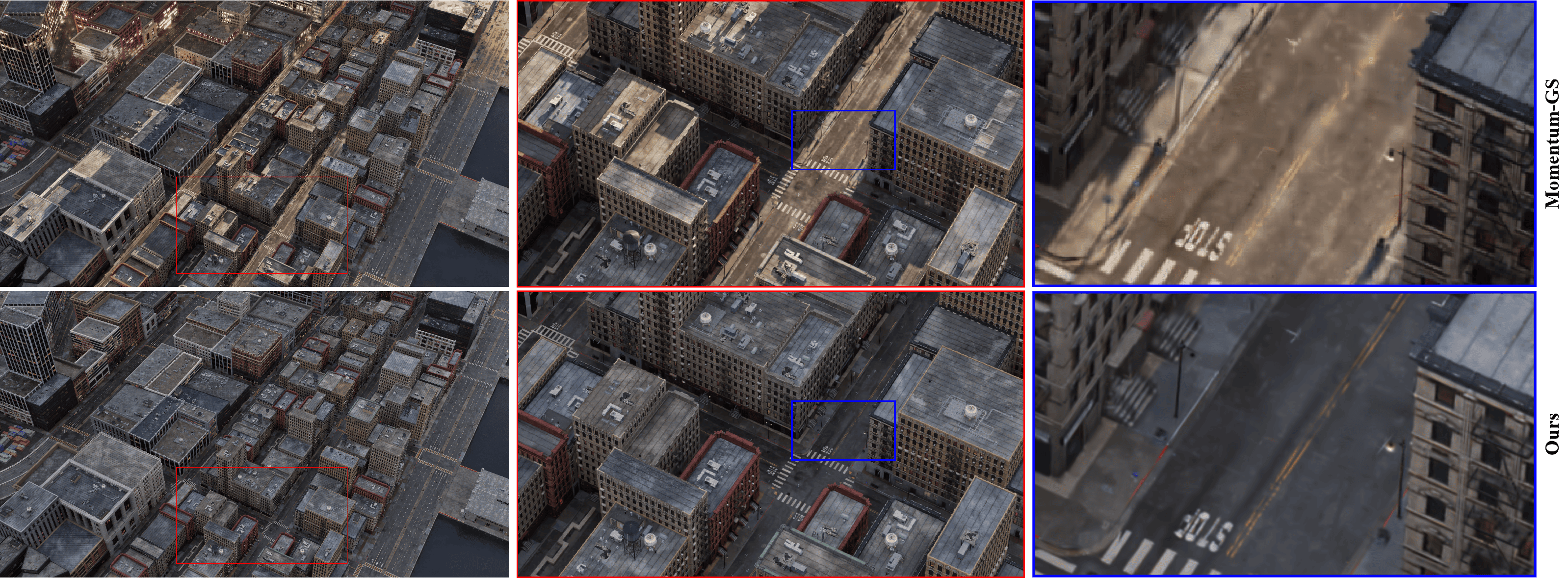}
\caption{Comparison of large-scale scene reconstruction under block-wise lighting conditions. \textit{Top:} Results from Momentum-GS, exhibiting noticeable inconsistencies and deteriorated reconstruction quality in finer details upon magnification. \textit{Bottom:} Results from our method, maintaining a more consistent overall appearance across the large scene and demonstrating superior detail reconstruction upon closer inspection.}
    \label{fig:big_scene}
\end{figure*}

\textit{i.} The ``w/o partition" results demonstrate that SMW-GS maintains robust appearance disentanglement even under large-scale and inconsistent visual conditions, outperforming existing in-the-wild methods across all metrics. However, geometric quality suffers. For example, on \textit{Block\_A$^\ast$}, SSIM drops by 0.081 and PSNR by 1.91 dB compared to the optimal ``\{3, 2\}" partition. This highlights the importance of spatial partitioning for maintaining reconstruction quality and ensuring sufficient supervision.

\textit{ii.} The results from different partitioning strategies, \textit{i.e.} ``w/o partition", ``\{2, 2\}", ``\{3, 2\}", and ``\{4, 2\}", reveal a clear trend: partitioning the scene into more blocks generally improves reconstruction quality, especially in terms of SSIM, as finer partitions help the model better capture local structures.
However, when the number of blocks exceeds a certain value, the performance gains become marginal. We hypothesize this is due to a trade-off between spatial decomposition and data allocation.
In our experiments, performance saturates at a ``\{3, 2\}" partition, where Gaussians are already well-optimized. Further partitioning does not significantly improve results, potentially introducing instability or fluctuations due to reduced data overlap and weaker global coherence.

\textit{iii.} The ``w/o PSG", ``w/o PSG-S1", and ``w/o PSG-S2" variants are designed to evaluate the effectiveness of our camera selection strategy in a detailed manner.
Compared to CityGaussian's method, our PSG strategy significantly improves reconstruction. Stage 1 is especially critical, which ensures full supervision coverage, particularly near block boundaries as shown in Fig. \ref{fig:supervised-difference}. The right figure demonstrates that the full PSG strategy provides enhanced supervision for points near block boundaries and corners, compared to the ``w/o PSG-S1" variant. Stage 2 further refines view selection, and while less impactful individually, its removal still degrades performance. Together, the two stages offer complementary benefits for robust large-scale reconstruction.


\textit{iv.} When GPU resources are limited, omitting RBT leads to notable performance drops due to biased supervision. Without alternating across partitions, only a subset of images contributes to training, weakening appearance disentanglement and affecting reconstruction quality. RBT plays a critical role when computational resources are limited, ensuring more balanced and effective optimization.

\begin{figure*}[ht] 
    \centering 
    \includegraphics[width=0.98\textwidth]{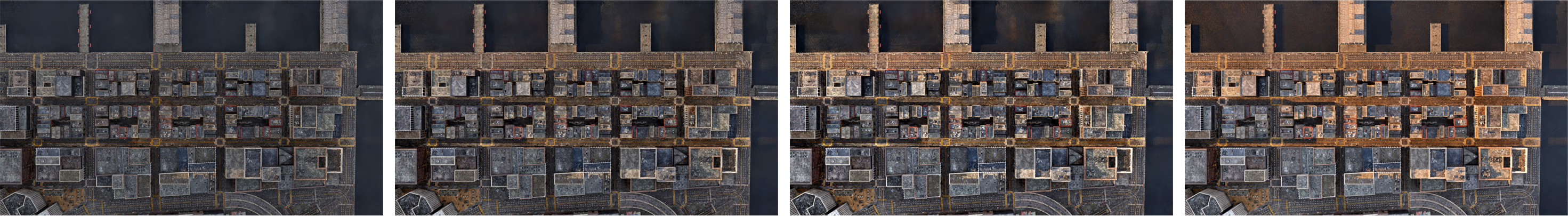}
    \caption{Appearance transition from daytime to dusk in a large-scale scene.}

    \label{fig:large_appearance_variation}
\end{figure*}

\begin{figure}[t]    
  \centering
  \includegraphics[width=0.95\columnwidth]{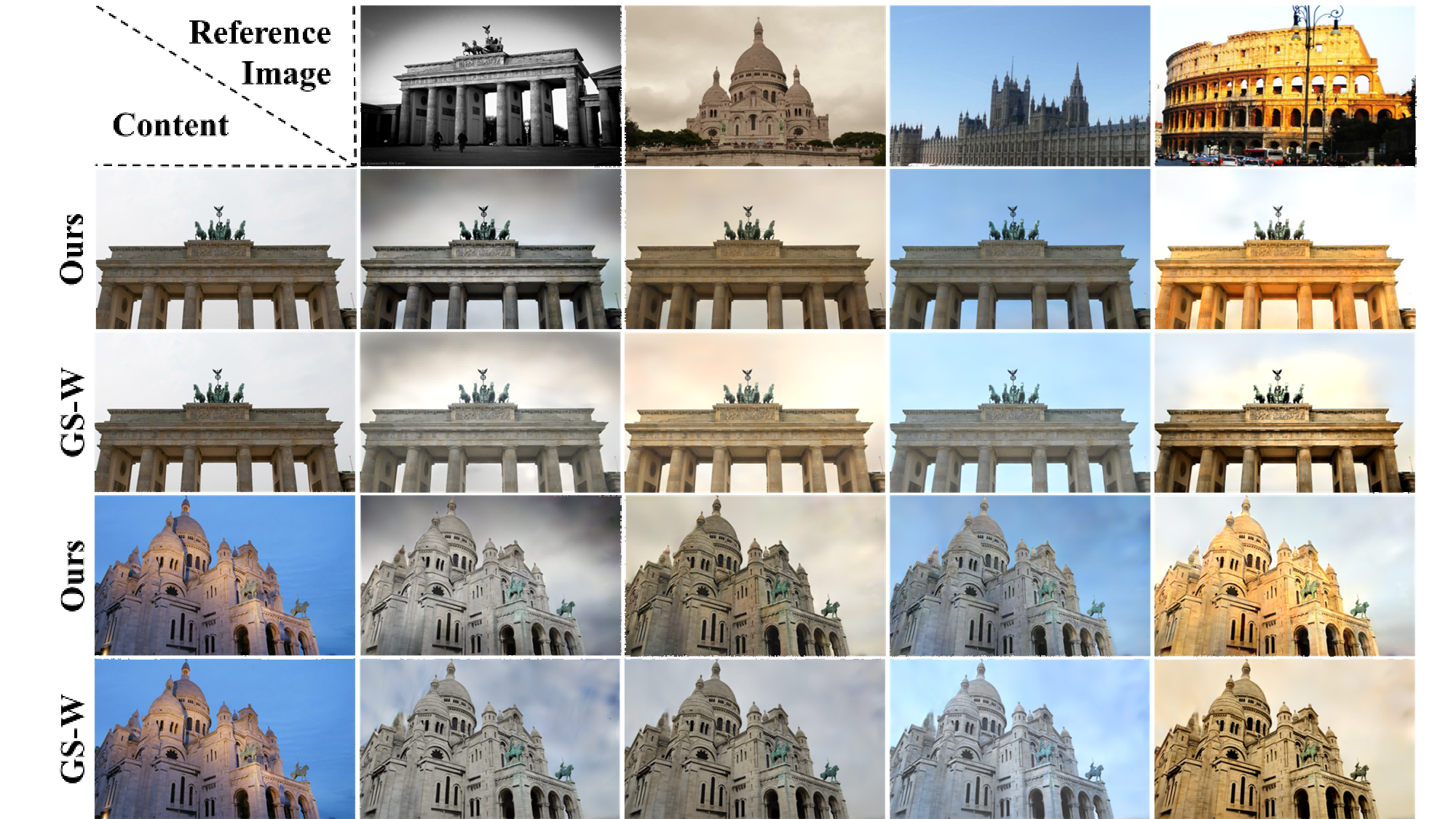}
  \caption{Qualitative comparison of appearance transfer performance across the \textit{Brandenburg Gate} and  \textit{Sacre Coeur} datasets.}
  \label{fig:transfer}
\end{figure}

\subsection{Extended Appearance Analysis}
To thoroughly evaluate our method's capability in handling complex real-world conditions, we conduct comprehensive experiments focusing on two critical aspects: (1) robustness under challenging illumination variations in large-scale environments, and (2) flexible appearance manipulation enabled by effective appearance component decoupling. These experiments validate both the reconstruction stability and the practical utility of our decomposed representation.

\subsubsection{Large-Scale Scene Robustness under Illumination Variations}
To simulate real-world lighting variations where illumination may remain consistent within local regions but varies across a large scene, we divide the scene into six spatial blocks, each using images with consistent intra-block but varying inter-block illumination (e.g., morning vs. evening captures). As shown in Fig. \ref{fig:large_appearance}, our method is compared to the recent SOTA method Momentum-GS under these challenging conditions.
As illustrated, Momentum-GS struggles to maintain consistent appearance across blocks, often showing stark appearance differences between adjacent regions (e.g., day-to-night shifts). In contrast, our method employs a globally trained appearance-extraction U-Net effectively disentangles complex lighting across the scene, enabling the reconstruction of scenes with consistent appearance across large-scale environments.

Fig. \ref{fig:big_scene} further highlights consistency and details. The top row illustrates results from Momentum-GS, while the bottom row shows those from our approach. Momentum-GS exhibits significant appearance inconsistencies across different image regions and suffers from blurred reconstruction details. In contrast, our method achieves superior visual coherence and sharper reconstruction quality.

Additionally, Fig. \ref{fig:large_appearance_variation} demonstrates our method's ability to manipulate the appearance of large-scale scenes (e.g., day to dusk) at a scene level. Unlike simple global color adjustments, our structured disentanglement supports region-specific appearance changes, corresponding to their distinct intrinsic properties. For instance, building areas exhibit subtle variations, whereas street regions experience more pronounced changes. 
\subsubsection{Appearance Transfer}  
Our method exhibits advanced capabilities for appearance transfer in 3D scenes, highlighting its robust and precise appearance modeling. As shown in Fig. \ref{fig:transfer}, a qualitative comparison between GS-W and our approach demonstrates that our method not only transfers both foreground and background elements to novel views but also retains intricate scene details, rather than merely reproducing the overall scene tone. This underscores the accuracy and reliability of our appearance modeling.


Overall, our method enables consistent, high-fidelity reconstructions and flexible appearance editing across large-scale scenes with complex lighting variations.

\section{Discussion and Conclusion}
\label{sec:conclusion}
This paper presents SMW-GS, a scalable 3D reconstruction framework that effectively handles unconstrained scenes with varying illuminatio. The key technical contributions include: (1) a triple-component appearance decomposition (global, refined, intrinsic) enabling explicit modeling of scene properties, (2) Micro-macro Wavelet Sampling for multi-scale feature extraction while preserving frequency-domain characteristics, and (3) a visibility-aware camera partitioning strategy that ensures consistent supervision across large-scale environments. Extensive experimental validation demonstrates that SMW-GS achieves superior reconstruction quality compared to state-of-the-art methods, particularly in challenging scenarios with complex illumination variations and large spatial extent.

Most existing methods, including SMW-GS, primarily focus on appearance variation, while transient occlusions are typically addressed using heuristic strategies limited to small-scale scenes. Effectively handling transient occluders in large-scale, sparsely-viewed environments remains a largely unsolved problem. Incorporating pretrained vision foundation models may offer a promising direction for addressing this challenge in future work. Additionally, although our method achieves efficient storage, its training and rendering remain slower than state-of-the-art methods due to appearance disentanglement and per-Gaussian color prediction. Caching precomputed appearance features could significantly improve rendering speed and is a key direction for future work.



 
%

\bibliography{smwgs}

\bibliographystyle{IEEEtran}

\end{document}